\pdfoutput=1
\documentclass[american,ngerman,english,twocolumn]{article}
\usepackage[T1]{fontenc}
\usepackage[latin9]{inputenc}
\usepackage{varioref}
\usepackage{textcomp}
\usepackage{amsmath}
\usepackage{graphicx}
\usepackage{authblk}

\makeatletter

\providecommand{\tabularnewline}{\\}

\RequirePackage{fix-cm}
\usepackage{graphicx}

\title{Robust Image Segmentation in Low Depth Of Field Images}
\author[]{Franz Graf}
\author[]{Hans-Peter Kriegel}
\author[]{Michael Weiler}
\affil[]{Ludwig-Maximilians-Universitaet Muenchen, \\Oettingenstr. 67, 80538 Munich, Germany \\[graf,kriegel,weiler]@dbs.ifi.lmu.de}

\renewcommand{\fnum@figure}{Fig.~\thefigure} 
\@ifundefined{showcaptionsetup}{}{%
 \PassOptionsToPackage{caption=false}{subfig}}
\usepackage{subfig}
\makeatother

\usepackage{babel}

\begin{document}
\maketitle

\begin{abstract}
In photography, low depth of field (DOF) is an important technique
to emphasize the object of interest (OOI) within an image. Thus, low
DOF images are widely used in the application area of macro, portrait
or sports photography. When viewing a low DOF image, the viewer implicitly
concentrates on the regions that are sharper regions of the image
and thus segments the image into regions of interest and non regions
of interest which has a major impact on the perception of the image.
Thus, a robust algorithm for the fully automatic detection of the
OOI in low DOF images provides valuable information for subsequent
image processing and image retrieval. In this paper we propose a robust
and parameterless algorithm for the fully automatic segmentation of
low DOF images. We compare our method with three similar methods and
show the superior robustness even though our algorithm does not require
any parameters to be set by hand. The experiments are conducted on
a real world data set with high and low DOF images.
\end{abstract}

\section{Introduction}

In photography, low depth of field (DOF) is an important technique
to emphasize the object of interest (OOI) within an image. Low DOF
images are usually characterized by a certain region which is displayed
very sharp like the face of a person and blurry image regions which
are significantly before of behind the object of interest. 

Low DOF images are well known from sports, portrait or macro photography
where only a specific part of the image should attract most of the
users' attention. The OOI is thereby displayed sharp while other areas
like the background appears blurred, so that the viewer automatically
focuses on the sharp areas of the image. When viewing a low depth
of field image, the viewer implicitly segments the image into regions
of interest and regions of less interest (usually background). As
this implicit segmentation has major impact on the perception of the
image, this information is a valuable feature for the subsequent image
processing chain like an adaptive image compression \cite{kavitha2009lossy}
or image retrieval aspects such as the similarity of images which
can be considerably influenced by the image's DOF. Given for example
two images displaying a person in the sharp image region in front
of different, blurred backgrounds, people might judge both pictures
similar even though the blurred background differs. Although this
implicit segmentation is rather easy for a human viewer of the photo,
it is not an easy task for a completely unsupervised algorithm. This
can be explained by the fact that there is usually not a sharp edge
which divides the sharp OOI and blurred background. Depending on the
camera's setting, this transition can be very smooth so that it is
hard to distinguish where the OOI ends.

With the vastly growing market of consumer DSLRs or even new small
compact cameras like the Sony Cybershot which are explicitly being
advertised with the ability for low DOF photos, the amount of low
DOF photos also increased. This growing amount of low DOF images may
also provide new information for established search and retrieval
systems if they take the OOI into account when performing the similarity
search tasks. In order to profit from the low DOF information, search
engines and feature extraction algorithms need fully automatic and
robust image segmentation algorithms which can separate the OOI from
the rest of the image. For large search engines or image stock agencies,
such algorithms should also be independent of the image domain, image
size and the color depth of the image so that the algorithm performs
well, no matter of the color or the toning of the photo (e.g. black-and-white,
color photos, sepia photos).

In this paper, we propose a robust, fully automatic and parameterless
algorithm for the segmentation of low DOF images as well as an analysis
of the impact of low DOF information on similarity search. The algorithm
does not need any a priory knowledge like image domain or camera settings.
The algorithm also provides meaningfull results even if the DOF is
rather large so that the background provides significant structures.
The rest of the paper is organized as follows: In Sec. \ref{sec:Related-work}
we review some related work and some technical background, follwed
by the explanation of the algorithm in Sec. \ref{sec:Algorithm}.
In Sec. \ref{sec:Experimental-Results} we explain our experimental
evaluation of the algorithm. In Sec. \ref{sec:Parameter-Settings}
we describe the \emph{internal} parameter settings and threshold values.
The impact of the DOF segmentation to image similarity is shown in
Sec. \ref{sec:Similarity}. Afterwards we finish the paper with a
conclusion and outlook in Sec. \ref{sec:Conclusion}.

\section{Related work\label{sec:Related-work}}

The segmentation of low DOF images has gained some interest in the
research community in past years. In \cite{li2002multiresolution,tsai1998segmenting}
early approaches to segment low DOF images were presented. Thereby
\cite{tsai1998segmenting} is using an edge-based algorithm which
first converts a gray-scale image into an edge-representation which
is then filtered. Afterwards the edges are linked to form closed boundaries.
These boundaries are treated with a region filling process, generating
the final result. \cite{li2002multiresolution} presents a fully automatic
segmentation algorithm using block based multiresolution wavelet transformations
on gray scale images. Even though the paper lists high rates of sensitivity
and specificity on the testset, the authors also name some limitations
like the dependence on very low DOF, fully focused OOI, and high image
resolution and quality. In \cite{ye2002unsupervised,won2002automatic}
high frequency wavelets are used to determine the segmentation of
low DOF images. As stated in \cite{kim2007fast}, these features have
the drawback of being not too robust if used alone and thus often
result in errors in both focused and defocused regions if the defocused
background shows some busy textures or if the focused foreground does
not have very significant textures. In \cite{kovacs2007focus}, localized
blind deconvolution is proposed to determine the focus map of an image.
Yet the authors do not propose a pure image segmentation algorithm
as the focus map is not a true segmentation but a measure for the
amount of focus in this part of the image. Also the algorithm does
not take into account any color information as it is only operating
on gray scale images. The works proposed in \cite{kim2007fast,park2006extracting,kim2005segmenting,kim2001video}
are consecutive works for segmentation of DOF images and sequences
of images \cite{kim2007fast,park2006extracting} like in movies which
address a similar topic. In this paper, we were inspired by the algorithm
proposed in \cite{kim2005segmenting} which uses morphological filtering
for the segmentation. Some problems of this algorithm were given by
background that showed significant structures or if a photo was taken
with high ISO values. Also the algorithm showed some problems if spatially
separated OOIs were shown in a single image. Another problem can be
raised by the size of the structuring element used in the algorithm
\cite[Sec. IV]{kim2005segmenting}.

We compare our algorithm with the work of \cite{kim2005segmenting},
with \cite{li2007unsupervized} where single frames of videos are
processed into a saliency map which is processed by morphological
filters. The resulting tri-map is then used for error control and
for the extraction of boundaries of the focused regions. We also compare
our work to the algorithm proposed in \cite{zhang-fuzzy}, where a
fuzzy segmentation approach was proposed by first separating the image
into regions using a mean shift. These regions are then characterized
by color features and wavelet modulus maxima edge point densities.
Finally, the region of interest and the background are separated by
defuzzification on fuzzy sets generated in the previous step. Our
test image dataset consists of a set of various photos and comprises
several categories from high to low DOF images.

\subsection{Depth of Field}

In optics, the DOF denotes the depth of the sharp area around the
focal point of a lens seen from the photographer. Technically, each
lens can only focus at a certain distance at a time. This distance
builds the focal plane which is orthogonal to the photographers view
through the lens. Precisely, only objects directly on the focal plane
are absolutely sharp, while objects before or behind the focal plane
are displayed unsharp. With increasing distance from the focal plane,
the sharpness of the displayed object decreases. Nevertheless, there
is a certain range before and behind the focal plane where objects
are recognized as sharp until a blur is perceived. The depth of this
region is then called the DOF. As the sharpness decreases gradually
with increasing distance from the focal plane, it is hard to determine
an exact range for the DOF as the limits of the sharp area are only
defined by the perceived sharpness. 

Points in the defocused areas appear blurred to a certain degree.
This is often modeled by a Gaussian kernel $G_{\sigma}$ as in Eq.~\ref{eq:gaussBlur}
where $\sigma$ denotes the spread parameter which affects the strength
of the blur. For a given image \emph{I}, the blurred representation
can then be created by a convolution $G_{\sigma}*I$.\begin{equation}
G_{\sigma}\left(x,y\right)=\frac{1}{2\pi\sigma^{2}}exp\left(-\frac{x^{2}+y^{2}}{2\sigma^{2}}\right)\label{eq:gaussBlur}\end{equation}

\begin{figure}
\begin{centering}
\includegraphics[width=0.95\columnwidth]{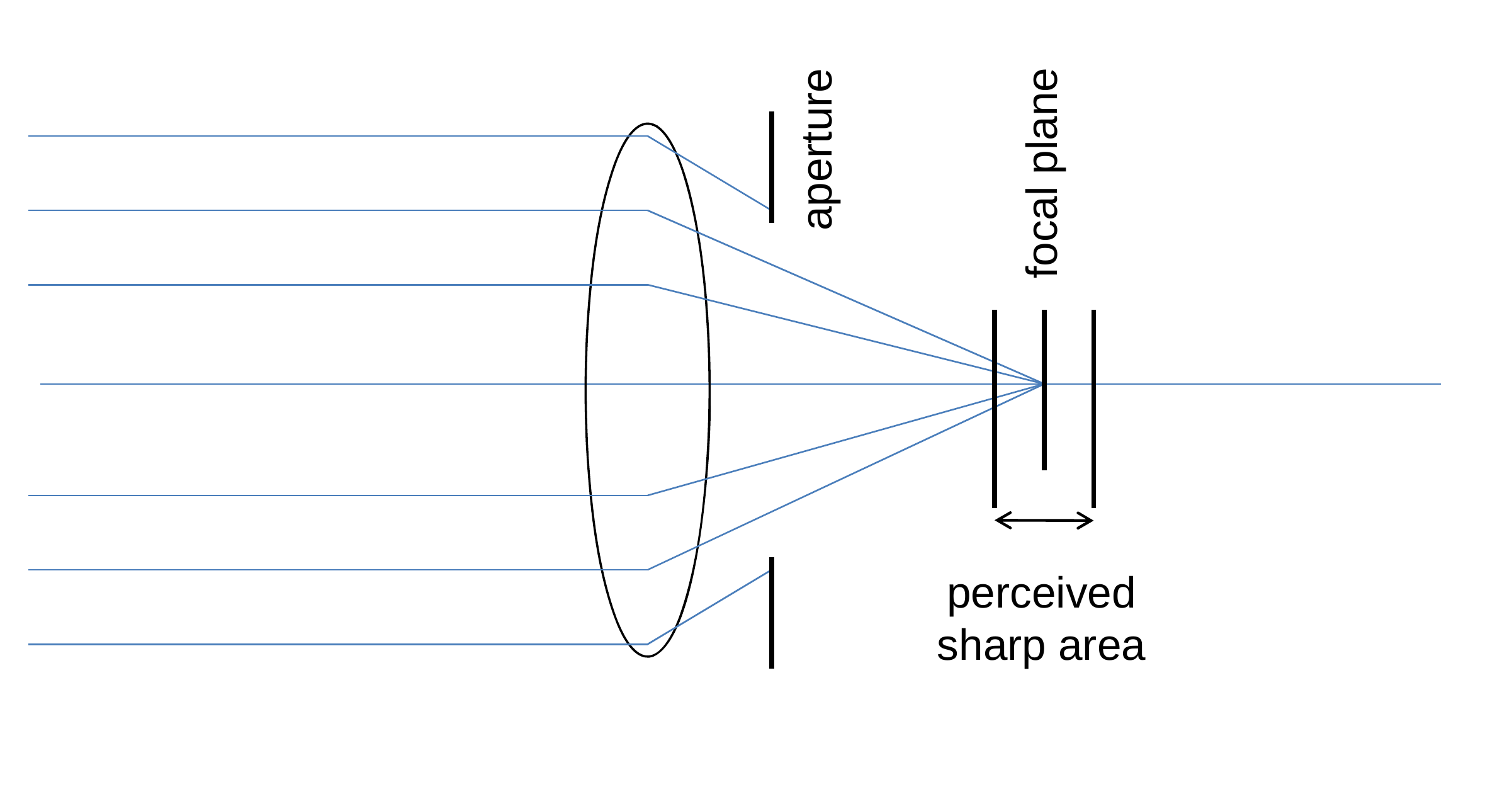}
\par\end{centering}

\caption{\label{fig:dof}Figure illustrating the depth of field. The size of
the perceived sharp area around the focal plane denotes the DOF.}

\end{figure}

The effect of DOF is mainly determined by the choice of the camera
respectively its imaging sensor size, aperture and distance to the
focussed object. The larger the sensor or aperture, the smaller the
DOF. Increasing the distance from the camera to the focussed object
will also expand the resulting DOF. Figure \ref{fig:dof} illustrates
the geometry of DOF at a symmetrical lens. 

\subsection{Automatic segmentation of low DOF images}

Automatic segmentation of images is more challenging than interactive
approaches because no additional information of humans can be used
to adapt parameter values for the segmentation process. However the
advantages of a fully automated algorithm are obvious, if the according
algorithm should be deployed to a system providing lots of images
where the segmentation should be present as fast as possible. This
is for example the case in search index or photo communities like
Flickr or Google's Picasa, where several thousand photos are uploaded
each minute, even if not all of them are low DOF images. 

The requirements to a segmentation algorithm are that it should be
able to handle different types (grayscale or color), orientations
(landscape or portrait) and resolutions (from small to large) of images,
independent of the camera settings like ISO etc. Many automatic segmentation
approaches of low DOF images have some of these restrictions, as seen
in \cite{zhang-fuzzy}, which only performs well on color images.
Grayscale images mostly fail because the extracted color features
are too few, to characterize regions and distinguish them sufficient.

However, other algorithms like the one presented in \cite{kim2005segmenting},
can only process grayscale images. In such cases, color images have
to be transformed and hence their color information looses its contribution
to improve segmentation quality. As shown in our experimental results
in section \ref{sec:Experimental-Results}, images that consist of
complex defocused regions can cause poor segmentation results, because
too many false positives are found. In this context, false positives\emph{
}describe the set of pixels that are defined as background by the
underlying ground truth but classified as OOI pixel by the segmentation
algorithm.

In the following section, we describe our algorithm that does not
suffer from one of the restrictions mentioned above. Therefore, we
use a robust method for calculating the amount of sharpness of a pixel
in relation to its neighbors by taking advantage of the $L^{*}a^{*}b^{*}$
color space, which offers a more accurate matching between numerical
and visual perception differences between colors. The $L^{*}a^{*}b^{*}$
model was favored over the well known $RGB$ and $CMYK$ color spaces,
as the $L^{*}a^{*}b^{*}$ model is designed to approximate human vision
better than the other color spaces.

To accomplish the problems caused by images consisting of numerous
less blurred pixel regions showing complex structures, we apply a
density-based clustering algorithm to all found sharp pixels. This
enables our algorithm to distinguish between sharp pixels belonging
to the main focus region of the OOI (if these pixels belong to the
largest found cluster) and noise pixels located in background structures.

\section{Algorithm\label{sec:Algorithm}}

\begin{figure*}
\subfloat{\includegraphics[width=0.155\textwidth]{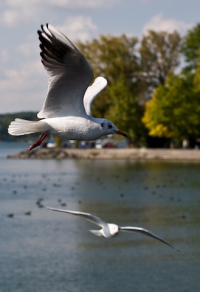}}\hfill{}\subfloat{\includegraphics[width=0.155\textwidth]{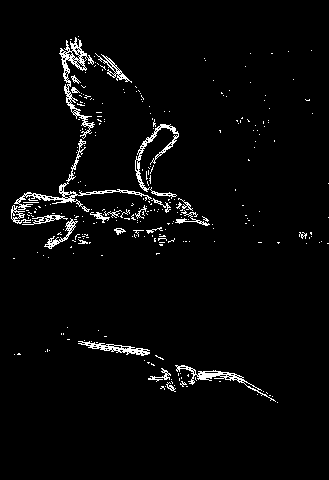}}\hfill{}\subfloat{\includegraphics[width=0.155\textwidth]{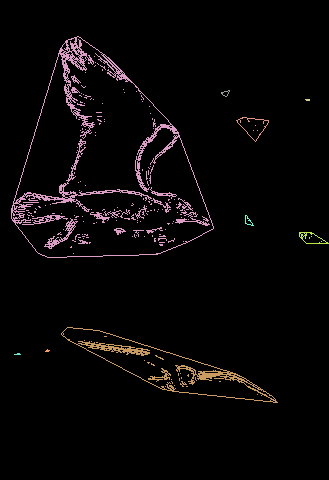}}\hfill{}\subfloat{\includegraphics[width=0.155\textwidth]{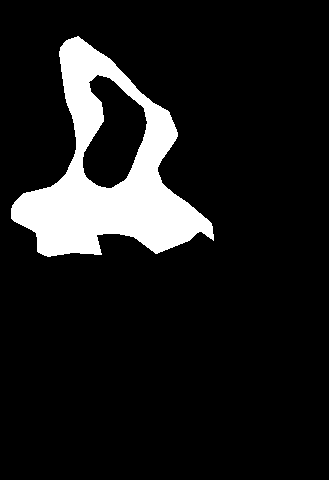}}\hfill{}\subfloat{\includegraphics[width=0.155\textwidth]{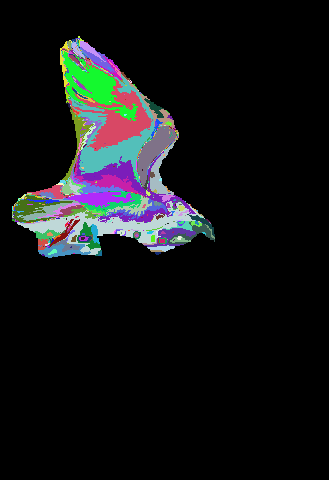}}\hfill{}\subfloat{\includegraphics[width=0.155\textwidth]{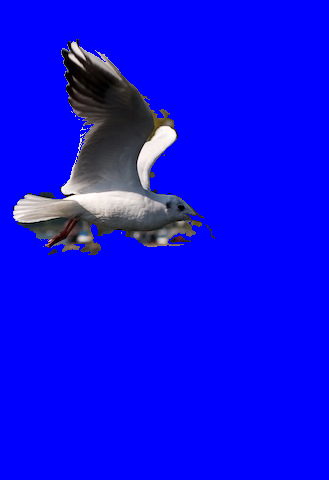}}\hfill{}\subfloat{\includegraphics[width=0.155\textwidth]{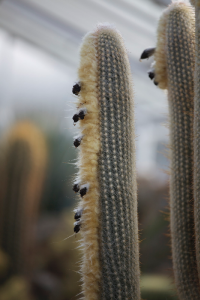}}\hfill{}\subfloat{\includegraphics[width=0.155\textwidth]{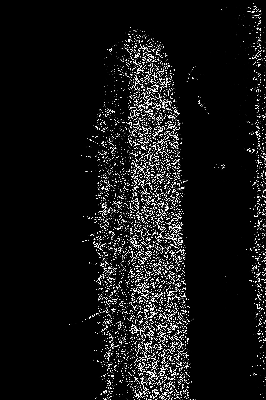}}\hfill{}\subfloat{\includegraphics[width=0.155\textwidth]{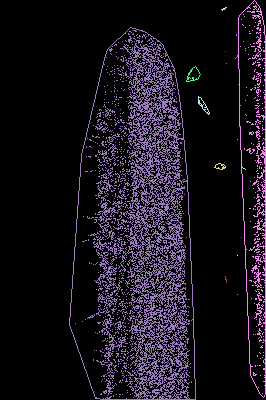}}\hfill{}\subfloat{\includegraphics[width=0.155\textwidth]{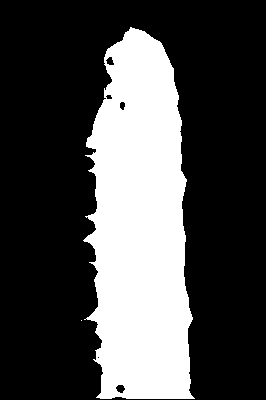}}\hfill{}\subfloat{\includegraphics[width=0.155\textwidth]{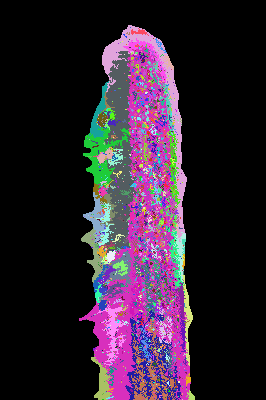}}\hfill{}\subfloat{\includegraphics[width=0.155\textwidth]{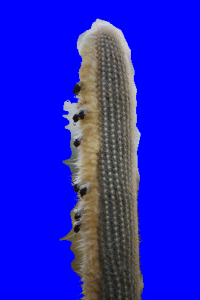}}\hfill{}\subfloat{\includegraphics[width=0.155\textwidth]{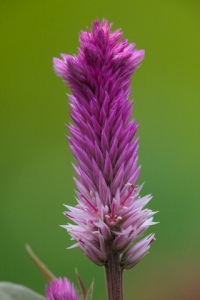}}\hfill{}\subfloat{\includegraphics[width=0.155\textwidth]{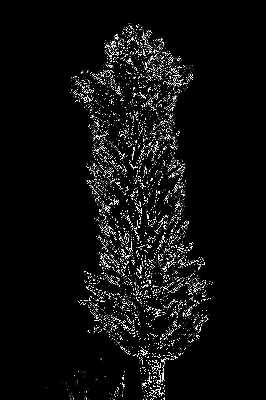}}\hfill{}\subfloat{\includegraphics[width=0.155\textwidth]{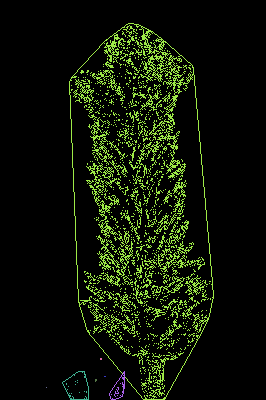}}\hfill{}\subfloat{\includegraphics[width=0.155\textwidth]{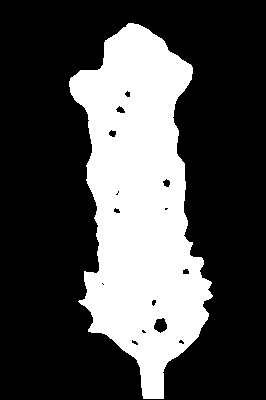}}\hfill{}\subfloat{\includegraphics[width=0.155\textwidth]{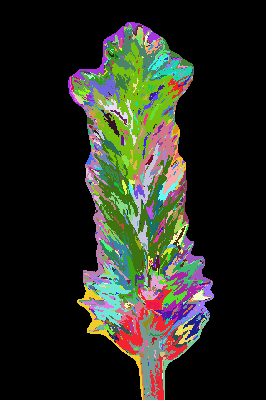}}\hfill{}\subfloat{\includegraphics[width=0.155\textwidth]{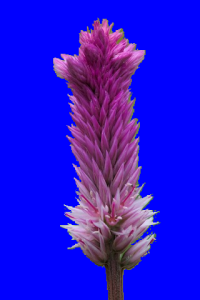}}\hfill{}\setcounter{subfigure}{0}\subfloat[\label{fig:Result-original-1}Input Image]{\includegraphics[width=0.155\textwidth]{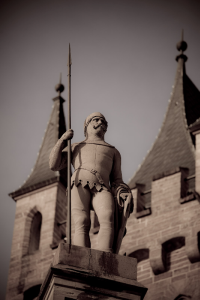}

}\hfill{}\subfloat[\textbf{\label{fig:Deviation-Scoring}}Deviation Scoring]{\includegraphics[width=0.155\textwidth]{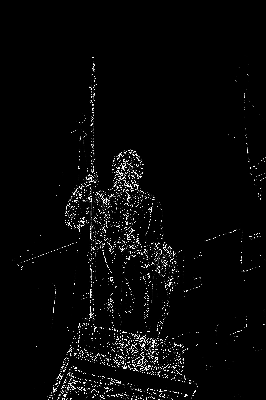}

}\hfill{}\subfloat[\textbf{\label{fig:Score-Clustering}}Score Clustering]{\includegraphics[width=0.155\textwidth]{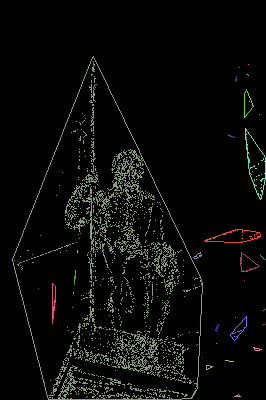}

}\hfill{}\subfloat[\textbf{\label{fig:Mask-Approximation}}Mask Approximation]{\includegraphics[width=0.155\textwidth]{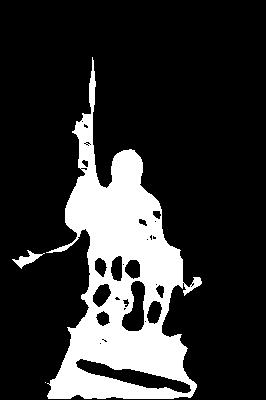}

}\hfill{}\subfloat[\textbf{\label{fig:Color-Segmentation}}Color Segmentation]{\includegraphics[width=0.155\linewidth]{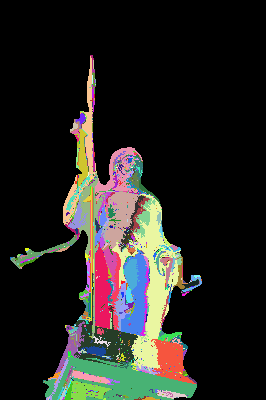}

}\hfill{}\subfloat[\textbf{\label{fig:Region-Scoring}}Region Scoring]{\includegraphics[width=0.155\textwidth]{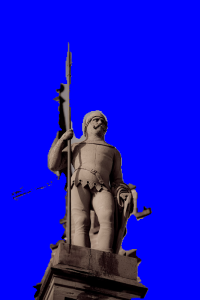}

}\caption{\label{fig:Results-1}Illustration of the five stages of our algorithm:
Fig.~\ref{fig:Result-original-1}: Input image with low DOF and relatively
complex background regions. Fig.~\ref{fig:Deviation-Scoring}: Identify
sharp pixels by computing the difference between the edges of the
original image and the edges of the blurred version of the image.
Fig.~\ref{fig:Score-Clustering}: Generate clusters from pixels with
a high appropriate score by a density-based clustering algorithm (for
a better visual representation we colored each found cluster and surround
it with its convex hull). Fig.~\ref{fig:Mask-Approximation}: Filling
all convex hulls from all neighbors of all dense pixels. Fig.~\ref{fig:Color-Segmentation}:
Group pixels into regions that contain similar colored pixels in the
original image. (For a better visual representation we colored each
found region in a random color). Fig.~\ref{fig:Region-Scoring}:
Removing all color Regions with low relevancy.}

\end{figure*}

The proposed algorithm consists of the following five stages: \emph{Deviation
Scoring}, \emph{Score Clustering}, \emph{Mask Approximation}, \emph{Color
Segmentation} and \emph{Region Scoring}. Before explaining the steps
of the algorithm in detail, we first want to give a brief summary
of the complete algorithm. Fig.~\ref{fig:Results-1} illustrates
the steps of the algorithm.

The first stage of the algorithm, called \emph{Deviation Scoring},
identifies sharp pixel areas in the image. Therefore a Gaussian Blur
is applied to the original image. The difference between the extracted
edges from the original image and the blurred image is then calculated.
For each pixel, this difference represents a score value, with higher
score values indicating sharper pixels and lower score values indicating
blurred pixels.

In the second stage, called \emph{Score Clustering}, all pixels with
a score value above a certain threshold are clustered by using a density-based
clustering algorithm. Thus, isolated sharp pixels are recognized as
noise and only large clusters are processed further.

The third stage named \emph{Mask Approximation} generates a nearly
closed plane (containing almost no holes) from the discrete points
of each remaining cluster. This is achieved by computing the convex
hull from all neighbors of all dense pixels. Any so-created polygon
is then filled and the union of these filled regions represent the
approximate mask of the main focus region. In the next two stages
this approximate mask is going to be refined.

Hence, the fourth stage, called \emph{Color Segmentation} divides
the approximate mask into regions that contain pixels with similar
color in the original image.

In the fifth stage named \emph{Region Scoring}, a relevance value
is calculated for each region. This relevance value is directly influenced
by the score values of the pixels surrounding the according region.
The final segmentation mask is then created by removing all regions
that have a relevance value below a certain threshold.%

\subsection{Deviation Scoring}

\begin{figure}
\subfloat[Original image]{\includegraphics[width=0.45\columnwidth]{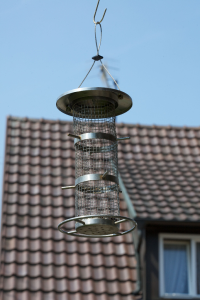}}\hfill{}\subfloat[Canny Edge Detection\label{fig:Canny-Edge-Detection}]{\includegraphics[width=0.45\columnwidth]{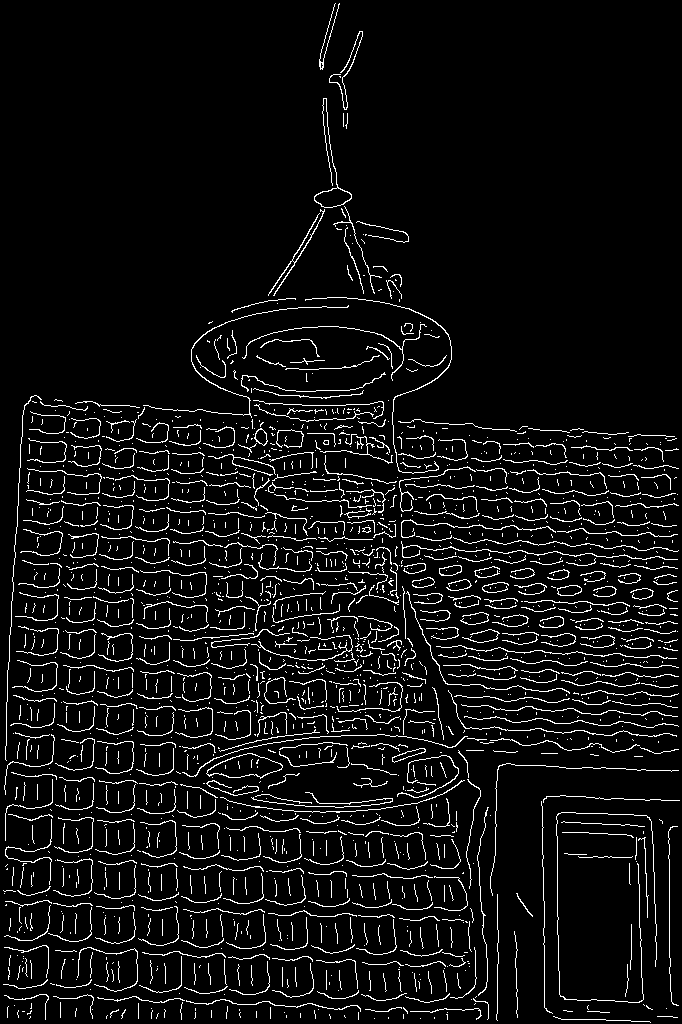}}

\subfloat[Higher Order Statistics\label{fig:Higher-Order-Statistics}]{\includegraphics[width=0.45\columnwidth]{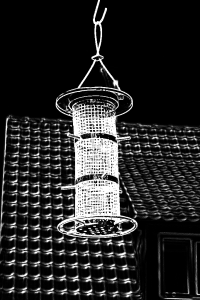}}\hfill{}\subfloat[Deviation Scoring\label{fig:Deviation-Scoring-1}]{\includegraphics[width=0.45\columnwidth]{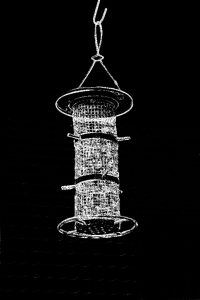}

}\caption{Comparison of edge detection techniques.}

\end{figure}

In the first step, we need to identify sharp pixels as an indication
for the focused objects within the image. The well known Canny edge
detector \cite{canny1987computational} is not suitable in this case
because the Canny detector operates on gray scale images and not on
the $L^{*}a^{*}b^{*}$ color space. Furthermore, the Canny operator
does not aim at the detection of single edge pixels but at the robust
detection of lines of edges even in partly blurred areas of the image
(c.f. Fig.~\vref{fig:Canny-Edge-Detection}). 

The HOS map used in \cite{kim2005segmenting} is defined as in Eq.
\ref{eq:HOS_map}\begin{equation}
HOS\left(x,y\right)=min\left(255,\frac{\hat{m}^{(4)}\left(x,y\right)}{DSF}\right),\label{eq:HOS_map}\end{equation}
where $DSF$ represents a down scaling factor of $100$ and the forth-order
moment $\hat{m}^{\left(4\right)}$ at $\left(x,y\right)$ is given
by $\hat{m}^{\left(4\right)}\left(x,y\right)=\frac{1}{N_{\eta}}\underset{\left(s,t\right)\in\eta\left(x,y\right)}{\overset{}{\sum}}\left(I\left(s,t\right)-\hat{m}\left(x,y\right)\right)^{4}$
where $\hat{m}$ is the sample mean and defined as in Eq. \ref{eq:mDach}
\begin{equation}
\hat{m}\left(x,y\right)=\frac{1}{N_{\eta}}\underset{\left(s,t\right)\in\eta\left(x,y\right)}{\overset{}{\sum}}I\left(s,t\right).\label{eq:mDach}\end{equation}
 Thereby $\eta\left(x,y\right)$ is the set of neighborhood pixels
with center $\left(x,y\right)$ and is set to size $3\times3$ where
$N_{\eta}$ denotes its cardinality. Using the HOS map also has the
disadvantage that it operates only on gray scale images. Additionally,
the HOS map is too sensitive in case of textured background as it
only produces reasonable results if the background is significantly
blurred. This works for images with very low DOF, but as soon as the
DOF is not very small, the HOS map detects too many sharp areas in
the background (c.f. Fig.~\vref{fig:Higher-Order-Statistics}).

\begin{figure}
\subfloat[\label{fig:deltaE_H}$HSV$ $\left(0,1,1\right)$ increased in component
$H$.]{\includegraphics[width=0.95\columnwidth]{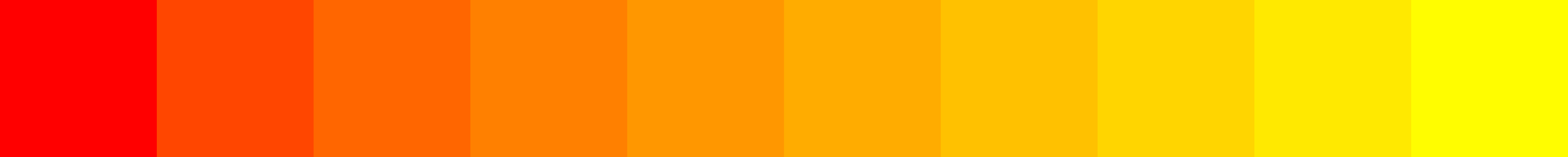}

}\hfill{}\subfloat[\label{fig:deltaE_S}$HSV$ $\left(0,0,1\right)$ increased in component
$S$.]{\includegraphics[width=0.95\columnwidth]{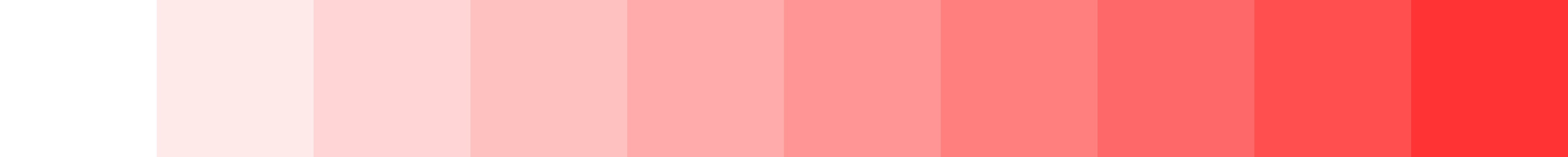}

}\hfill{}\subfloat[\label{fig:deltaE_V}$HSV$ $\left(0,0,0\right)$ increased in component
$V$.]{\includegraphics[width=0.95\columnwidth]{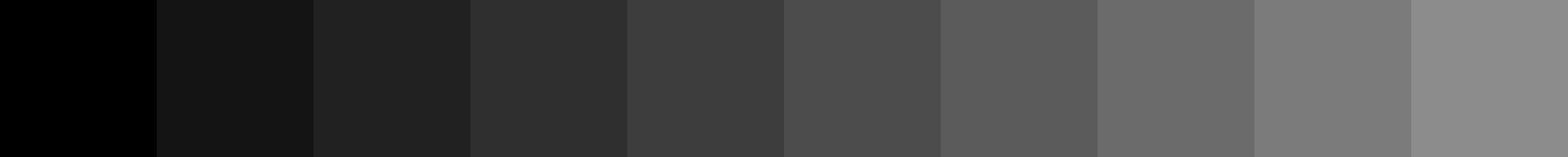}

}\caption{\label{fig:deltaEDistances}Visualization of the $\Delta E^{*}$ color
distance within each component of the well known $HSV$ color space,
with colors $c_{0},\ldots,c_{9}$. Where $c_{i}$ of the \emph{i}-th
square is increased in each of the components $H$ (Fig.~\ref{fig:deltaE_H}),
$S$ (Fig.~\ref{fig:deltaE_S}) and $V$ (Fig.~\ref{fig:deltaE_V})
so that $\Delta E^{*}\left(c_{i},c_{i+1}\right)\approx16$.}

\end{figure}

Thus, we propose the process of Deviation Scoring. Let \emph{$I$}
be the set of pixels of the processed image. For each pixel $p\left(x,y\right)\in I$,
the mean color from the pixel's $r$-neighborhood is calculated by\[
\eta_{I\left(x,y\right)}^{r}=\{p\left(x',y'\right)\mid\left|x'-x\right|\leq r\wedge\left|y'-y\right|\leq r\}\]
with $r$\emph{ }representing the L1-distance to the pixel $p(x,y)$.
The color value of $p(x,y)$ is represented in the $L^{*}a^{*}b^{*}$
color space and denoted by $\left(L_{p}^{*},A_{p}^{*},B_{p}^{*}\right)$.
Thus, the mean neighborhood color of $p(x,y)$ in the $L^{*}$-band
is determined by\[
L_{\eta_{I\left(x,y\right)}^{r}}=\frac{\underset{p\in\eta_{I\left(x,y\right)}^{r}}{\sum}\left(L_{p}^{*}\right)}{\left|\eta_{I\left(x,y\right)}^{r}\right|}.\]
The values for the $a^{*}$- and $b^{*}$-band are denoted by $a_{\eta_{I\left(x,y\right)}^{r}}$
and $b_{\eta_{I\left(x,y\right)}^{r}}$ respectively, so that the
mean neighborhood color $Lab_{\eta_{I\left(x,y\right)}^{r}}$of a
pixel $p(x,y)$ is defined by $\left(L_{\eta_{I\left(x,y\right)}^{r}},a_{\eta_{I\left(x,y\right)}^{r}},b_{\eta_{I\left(x,y\right)}^{r}}\right)$.
According to the International Commission on Illumination\emph{ CIE}%
\footnote{CIE: Commission Internationale de l'eclairage, http://www.cie.co.at%
}, the color distance $\Delta E^{*}\left(u,v\right)$ between two color
values $u$, $v$ in the $L^{*}a^{*}b^{*}$ color space is calculated
by using the Euclidean distance:\[
\Delta E^{*}\left(u,v\right)=\sqrt{\left(L_{u}^{*}-L_{v}^{*}\right)^{2}+\left(a_{u}^{*}-a_{v}^{*}\right)^{2}+\left(b_{u}^{*}-b_{v}^{*}\right)^{2}}\]
For each $p\left(x,y\right)$, the neighbor difference $\Delta\eta_{\left(x,y\right)}^{r}$
is then defined by\[
\Delta\eta_{I\left(x,y\right)}^{r}=min\left(255\cdot\frac{\Delta E^{*}\left(Lab_{\eta_{I\left(x,y\right)}^{r}},p\right)}{\Delta E^{max}},\,255\right)\]
with $\Delta E^{max}$ being the maximum possible distance in the
$L^{*}a^{*}b^{*}$ color space and $\Delta E^{*}\left(u,v\right)$
being the Euclidean distance of the color values $u$, $v$ in the
$L^{*}a^{*}b^{*}$ space. In Fig. \ref{fig:deltaEDistances} we illustrated
a color distance of $\Delta E^{*}=16$ , by varying one of the three
components of a base color defined in the $HSV$ color space .

In the following, all pixels $p\left(x,y\right)$ with a neighbor
difference greater than the threshold $\Theta_{score}\in\left[0,\,255\right]$
are called \emph{edges} or \emph{edge pixels}, so that the equation
$\Delta\eta_{I\left(x,y\right)}^{r}>\Theta_{score}$ holds for each
edge pixel of the image. Even though the parameter $\Theta_{score}$
could be set freely, we recommend a value of 50 (c.f. Tab.~\ref{tab:Parameters})
as it showed the best result. Before calculating the score values
of the edge pixels, $I$ is convolved using a Gaussian kernel with
a standard deviation $\sigma=\Theta_{\sigma}$ to remove noise and
generally soften the image. We recommend to set the value of $\Theta_{\sigma}$
to $\frac{1}{10}$ (c.f. Tab.~\ref{tab:Parameters}). The resulting
image is then denoted by $I'$.

Afterwards, another image $I'_{\sigma}$ is created by convolving
$I'$ once again by using the same Gaussian kernel. $I'$and $I'_{\sigma}$are
then used to compute the score values $\mu\in\left[0,255\right]$
of the edge pixels. Therefore, the score $\mu(x,y)$ for an edge pixel
$p\left(x,y\right)$ is determined by the squared neighbor difference
in the images $I'$and $I'_{\sigma}$ at the location of the according
pixel:\[
\mu\left(x,y\right)=min\left\{ 255,\left(\Delta\eta_{I'\left(x,y\right)}^{r}-\Delta\eta_{I'_{\sigma}\left(x,y\right)}^{r}\right)^{2}\right\} \]
Due to the limitation to $\mu(x,y)\leq255$, we are treating all color
changes between $I'(x,y)$ and $I(x,y)$ equally where $\Delta E>16$.
This can be justified by human perception, which recognizes two colors
$u$, $v$ to as rather unsimilar to each other if $\Delta E^{*}(u,v)>12$
\cite{chung2001quantitative}. Thus it can be said, that a $\Delta E^{*}>16$
indicates a significant color change which is also a strong indication
for an edge.

Afterwards, all edge pixels with a score value greater than the threshold
$\Theta_{score}$ are treated as candidates for the focused region
of the image while the score values of all pixels having a score value
less than $\Theta_{score}$ are set to $0$ and are thus no candidates.
The resulting candidate set $I_{score}(x,y)$, is defined by the following
equation:\[
I_{score}\left(x,y\right)=\begin{cases}
0 & \mu\left(x,y\right)<\Theta_{score}\\
\mu\left(x,y\right) & else\end{cases}\]
An illustration of the candidate set can be seen in Fig.~\ref{fig:Deviation-Scoring-1}
and Fig.~\ref{fig:Deviation-Scoring}, where brighter pixels indicate
a large score and black pixels indicate a score less than the threshold
$\Theta_{score}$.

\subsection{Score Clustering\label{sub:Score-Clustering}}

In this step, clusters are generated from all points in $I_{score}$
in order to find compound regions of focused areas. Therefore, the
density-based clustering algorithm DBSCAN~\cite{ester1996density}
is used. In contrast to the K-Means~\cite{Har75}, which partitions
the image into convex clusters, DBSCAN also supports concave structures
which is more desirable in this case. The following section gives
a short outline of DBSCAN and then describes how the necessary parameters
$\varepsilon$ and $minPts$ are determined automatically and how
DBSCAN is used in our segmentation algorithm for further processing.

\subsubsection{DBSCAN}

In this stage, clusters are generated from all $p\in I_{score}$ by
applying DBSCAN, which is based on the two parameters $\varepsilon$
and $minPts$. The main idea of this clustering algorithm is that
each point in a cluster is located in a dense neighborhood of other
pixels belonging to the same cluster. The area in which the neighbors
must be located is called the \emph{$\varepsilon$-neighborhood} of
a point, denoted by $N_{\varepsilon}\left(p\right)$, which is defined
as follows:\begin{equation}
N_{\varepsilon}\left(p\right)=\left\{ q\in D\mid dist\left(p,q\right)\leq\varepsilon\right\} \end{equation}
where \emph{D} is the database of points and $dist\left(p,q\right)$
describes the distance measure (e.g. the Euclidean distance) between
two points $p,q\in D$. 

For each point \emph{$p$} of a cluster \emph{$C$} there must exist
a point \emph{$q\in C$} so that \emph{$p$} is within the \emph{$\varepsilon$-neighborhood}
of \emph{$q$} and $N_{\varepsilon}\left(p\right)$ includes at least
\emph{$minPts$} points. Therefore some definitions are invoked which
are described in the following. Considering \emph{$\varepsilon$}
and \emph{$minPts$}, a point \emph{$p$} is called directly \emph{density-reachable}
from another point \emph{$q$}, if $p\in N_{\varepsilon}\left(p\right)$
and $p$ is a so-called \emph{core-point.} A point $p$ defined as
a \emph{core-point} if $\left|N_{\varepsilon}\left(p\right)\right|\geq MinPts$
holds. If there exists a chain of \emph{$n$} points $p_{1},\,\ldots,\, p_{n}$,
such that $p_{i+1}$ is \emph{directly density-reachable} from $p_{i}$,
then $p_{n}$ is called \emph{density-reachable} from $p_{1}$. Two
points \emph{$p$} and \emph{$q$} are \emph{density-connected} if
there is a point \emph{$o$} from which \emph{$p$} and \emph{$q$}
are both density reachable, considering \emph{$\varepsilon$} and
\emph{$minPts$.}

Now, a cluster can be defined as a non-empty subset of the Database
\emph{$D$}, so that for each \emph{$p$} and \emph{$q$}, the following
two conditions hold\emph{:}
\begin{itemize}
\item $\forall p,\, q:$ if $p\in C$ and \emph{$q$} is \emph{density-reachable}
from \emph{$p$, }then\emph{ $q\in C$}
\item $\forall p,\, q\in C:$ \emph{$p$} is density-connected to \emph{$q$}
\end{itemize}
Points that do not belong to any cluster are treated as $noise=\left\{ p\in D\mid\forall i:p\notin C_{i}\right\} $,
where $i=1,\,\ldots,\, k$ and C$_{1},\,\ldots,\, C_{k}$ are the
found clusters in \emph{$D$}.

\subsubsection{Determination of Parameters\label{sub:Determination-of-Parameters}}

To provide highest flexibility with respect to the different occurrences
of the focused area, we do not apply absolute values for $\varepsilon$
and $minPts$, but compute them relatively to the size of the image
and its score distribution. Thus, $\varepsilon$ is calculated by
$\varepsilon=\sqrt{\left|I\right|}\cdot\Theta_{\varepsilon}$, with
$\left|I\right|$ denoting the total amount of pixels of the image
represented by \emph{$I$} and $\Theta_{\varepsilon}\in\left[0,1\right]$.
The second parameter $minPts$ is determined by\[
minPts=\left\lfloor \frac{\left(\varepsilon+1\right)^{2}}{|I|}\left(\underset{(x,y)\in I_{score}}{\sum}min\left\{ \frac{\mu(x,y)}{\Theta_{dbscan}},1\right\} \right)\right\rfloor ,\]
with the threshold $\Theta_{dbscan}$ set to $255$.

The result of the DBSCAN clustering is a cluster set $C=\left\{ c_{1},\ldots,\, c_{n}\right\} $,
with each $c_{i}\in C$ representing a subset of pixels $p(x,y)\in I$.
Due to our assumption that small isolated sharp areas are treated
as noise, we define the relevant score cluster set \[
\hat{C}=\left\{ c\in C\mid|c_{i}|\geq\frac{max_{C}}{2}\right\} \subseteq C,\]
 with $max_{C}=max\left\{ \left|c_{1}\right|,\ldots,\left|c_{n}\right|\right\} $
being the amount of pixels of the largest cluster. An illustration
of this step can be seen in Fig.~\vref{fig:Score-Clustering} where
different clusters are painted in different colors.

\subsection{Mask Approximation\label{sub:Mask-Approximation}}

The relevant score cluster set $\hat{C}$, as defined in section \ref{sub:Score-Clustering},
is already a good reference point of the OOI's location and distribution.
In general however, there exists no single contiguous area, but several
individual regions of interest representing the focused objects. This
stage of the algorithm connects all clusters $c\in\hat{C}$ to a contiguous
area which represents an approximate binary mask of the OOI. This
is achieved through the two steps \emph{Convex Hull Linking} and\emph{
Morphological Filtering, }that will be described in more detail below.

\subsubsection{Convex Hull Linking}

In the \emph{convex hull linking} step we first generate the convex
hull for all points in the \emph{$\varepsilon$-}neighborhood\emph{
}$N_{Eps}\left(p\right)$ of each core point \emph{$p$} of the cluster
set. Let $K=\left\{ k_{1},\ldots,k_{j}\right\} $ be the set of all
core points from the score clusters in $\hat{C}$ and let $convex(P)$
be the convex hull of a point set \emph{$P$. }Then we can define
the set of convex hull polygons by $H=\left\{ convex(N_{eps}\left(k_{1}\right)),\ldots,convex(N_{eps}\left(k_{j}\right))\right\} $
which is used to generate a contiguous area. Therefore each $p\left(u,v\right)\in I$
is checked, if it is located within one of the convex hull polygons
of \emph{$H$. }If that is the case, we mark this pixel with 1, otherwise
with 0. The binary approximation mask $I_{app}$ is then given by
\[
I_{app}\left(x,y\right)=\begin{cases}
1 & if\:\exists H_{i}:p(x,y)\in H_{i}\\
0 & otherwise\end{cases}.\]

Afterwards we apply the morphological filter operations \emph{closing}
and \emph{dilation by reconstruction} to $I_{app}$ for smoothing
and closing small holes.

\subsubsection{Morphological Filtering}

Morphological filters are based on the two primary operations \emph{dilation}
$\delta_{H}\left(I\right)$ and \emph{erosion} $\varepsilon_{H}\left(I\right)$\emph{
}where\emph{ $H\left(i,j\right)$ $\in\left\{ 0,1\right\} $} denoting
the \emph{structuring element. }For\emph{ }a binary image \emph{I},
$\delta_{H}\left(I\right)$ and $\varepsilon_{H}\left(I\right)$ are
defined as in the following equations:\[
\delta_{H}\left(I\right)=\left\{ \left(s,t\right)=\left(u+i,v+j\right)\mid\left(s,t\right)\in I,\left(i,j\right)\in H\right\} \]
\[
\varepsilon_{H}\left(I\right)=\left\{ \left(s,t\right)\mid\left(s+i,t+j\right)\in I,\forall\left(i,j\right)\in H\right\} \]
The operation \emph{morphological} \emph{closing} $\varphi_{H}$,\emph{
}is a composition from the two primary operations, so that $\varphi_{H}\left(I\right)=\varepsilon_{H}\left(\delta_{H}\left(I\right)\right)$.
Thus, the input image \emph{$I$} is initially dilated and subsequently
eroded, both times with the same \emph{structuring element} \emph{H}.
In order to define the following operation \emph{dilation by reconstruction,
}some more definitions are required. At first, the primary operations
$\delta_{H}\left(I\right)$ and $\varepsilon_{H}\left(I\right)$ are
extended to the\emph{ basic geodesic dilation} $\delta^{\left(1\right)}\left(I,I'\right)$
and \emph{basic geodesic} \emph{erosion} $\varepsilon^{\left(1\right)}\left(I,I'\right)$
of size one as in the following equations.\[
\delta^{\left(1\right)}\left(I,I'\right)\left(u,v\right)=min\left\{ \delta_{H}\left(I\right)\left(u,v\right),I'\left(u,v\right)\right\} \]
\[
\varepsilon^{\left(1\right)}\left(I,I'\right)\left(u,v\right)=max\left\{ \varepsilon_{H}\left(I\right)\left(u,v\right),I'\left(u,v\right)\right\} \]
Note that these basic geodesic operations need an additional Image
$I'$\emph{, }which is called marker, where the input image \emph{I}
is called mask. Thus, the result of a \emph{geodesic erosion} at position
$\left(u,v\right)$ is the maximum value of the \emph{erosion} $\varepsilon_{H}$
of mask \emph{I} and the value of the marker image $I'\left(u,v\right)$
and vice versa for the \emph{geodesic dilation}. The geodesic \emph{erosion}
$\varepsilon^{\left(\infty\right)}$ and \emph{dilation} $\delta^{\left(\infty\right)}$
of infinite size, called \emph{reconstruction by erosion $\varphi^{\left(rec\right)}$
}and\emph{ reconstruction by dilation} $\gamma^{\left(rec\right)}$,
is then defined as follows\[
\varphi^{\left(rec\right)}\left(I,I'\right)=\varepsilon^{\left(\infty\right)}\left(I,I'\right)=\varepsilon^{\left(1\right)}\circ\ldots\circ\varepsilon^{\left(1\right)}\left(I,I'\right)\]
\[
\gamma^{\left(rec\right)}\left(I,I'\right)=\delta^{\left(\infty\right)}\left(I,I'\right)=\delta^{\left(1\right)}\circ\ldots\circ\delta^{\left(1\right)}\left(I,I'\right)\]
Note that $\varphi^{\left(rec\right)}\left(\cdot,\cdot\right)$ and
$\gamma^{\left(rec\right)}\left(\cdot,\cdot\right)$ converge and
achieve stability after a certain number of iterations. Thus it is
assured that these functions do not need to be executed indefinitely
and so the application is guaranteed to terminate.

\subsubsection{Application}

In our approach, we primarily apply a \emph{morphological closing}
operation $\varphi_{H}\left(I_{app}\right)=\varepsilon_{H}\left(\delta_{H}\left(I_{app}\right)\right)$
to the approximate mask. The dimension of the structuring element
\emph{$H$} therefore is discussed later. Afterwards we use $\varphi^{rec}\left(I_{app},\delta_{H'}\left(I_{app}\right)\right)$
to close holes in the approximate mask $I_{app}$. The dimension of
the structuring element \emph{$H$} is $h\times h$, where \emph{h}
is calculated relatively to the total pixel count $\left|I_{app}\right|$
of the image \emph{$I_{app}$,} so that $h=\sqrt{\left|I_{app}\right|}\cdot\Theta_{rec}$,
with $\Theta_{rec}\in\left[0,1\right]$. After this morphological
processing, the approximate mask $I_{app}$ covers the OOI quite well
(c.f. Fig.~\vref{fig:Mask-Approximation}). In general however, it
includes boundary regions that exceed the borders of the OOI and tend
to surround it with a thick border. The following two stages of our
algorithm refine the mask by erasing the surrounding border regions.

\subsection{Color Segmentation}

In this stage, the pixels from the approximate mask $I_{app}$ are
divided into groups, so that each group contains pixels that correspond
to similar colors in \emph{$I$}. Therefore we process each $p\left(u,v\right)\in I_{app}$
and iteratively include all its neighbors $n$ for which the following
conditions hold:\[
n\in\left\{ \left(s,t\right)\in\eta_{I_{app}\left(u,v\right)}^{1}\mid I_{app}\left(s,t\right)=I_{app}\left(u,v\right)=1\right\} \]
\begin{equation}
\wedge\mbox{\,\,}\Delta E^{*}\left(p,n\right)<\Theta_{dist}.\label{eq:ColorSeg2}\end{equation}
The threshold $\Theta_{dist}\in\left[0,100\right]$ is an internal
parameter, which specifies the maximum distance between two color
values $u,v$ in the $L^{*}a^{*}b^{*}$ color space.

Therefore a method\emph{ $expand(x,y,R)$ }is called for each $p\left(x,y\right)\in\{\left(x,y\right)\in I_{app}\mid I_{app}\left(x,y\right)=1\}$,
which is not yet marked as visited. $R=\left\{ \left(x,y\right)\right\} $
here defines a new color region formed by the point $p_{1}\left(x,y\right)$.
The method \emph{$expand(x,y,R)$} then proceeds as follows: For all
neighbors $p_{2}(x,y)$ of $p(x,y)$ fulfilling Eq.~\vref{eq:ColorSeg2}
we add $p_{2}$ to \emph{$R$} and mark $p\left(u,v\right)$ as visited.
Then $expand\left(u,v,R\right)$ is called recursively. The resulting
set of regions is called $R_{color}$.

\subsection{Region Scoring\label{sub:Region-Scoring}}

In this step, a relevance value $\mu$ is calculated for each region
$r\in R_{color}$. The more a region is surrounded by areas with a
large score $\mu$, the larger the relevancy value gets. Low relevant
regions are removed afterwards which causes an update of $\mu$ in
the neighboring regions and thus possibly triggers another deletion
if the relevance of an updated region is not high enough after the
according update.

\subsubsection{Boundary Overlap}

The boundary overlap\emph{ }$BO_{r}^{R}$ of a region $r$ is a measure
for the adjacency of \emph{$r$} to the approximate mask $I_{app}$
and is defined as \[
BO_{r}^{R}=\left|\{\left(u,v\right)\in B_{r}\mid\exists r'\in R:\left(u,v\right)\in r'\}\right|,\]
where $B_{r}$ is the difference of \emph{$r$} to its dilation. The
mask boundary overlap $MBO_{r}$ of \emph{$r$} is then defined as
$MBO_{r}=\frac{BO_{r}^{R_{color}}}{\left|B_{r}\right|}.$ $MBO_{r}$
specifies the ratio of the number of outline points located in other
regions to the number of all outline points of \emph{$r$}. 

The score boundary overlap $SBO_{r}=\frac{BO_{r}^{\hat{C}}}{\left|B_{r}\right|}$
of \emph{$r$} is a measure for the adjacency of \emph{$r$} to the
corresponding score values $\mu$. A large $SBO_{r}$ indicates, that
\emph{$r$} has a neighborhood with large corresponding score values
$\mu$.

\subsubsection{Mask Relevance}

The mask relevance for a given region \emph{$r$ }can then be defined
as $MR_{r}=SBO_{r}\cdot MBO_{r}$. Afterwards, we eliminate all regions
\emph{$r$} with a mask relevance value which is too low. The calculation
of $MR_{r}$ is executed iteratively: Let $MR_{r}^{i}$ denote the
value $MR_{r}$ of a region \emph{$r$} during the \emph{$i$}-th
iteration. One iteration cycle computes the corresponding $\mu$ for
each region $r$ and deletes \emph{$r$} from the approximate mask
$I_{app}$ if $MR_{r}^{i}\leq\Theta_{rel}$. The precise assignment
of $\Theta_{rel}$ and its impact on segmentation quality is discussed
later. Once a region \emph{$r$} satisfies $MR_{r}^{i}\leq\Theta_{rel}$
at iteration \emph{$i$}, it will be erased from $I_{app}$, so that
$\forall\left(x,y\right)\in r:\, I_{app}\left(x,y\right)=0$. The
calculation of $MR_{r}^{i}$ continues for $i=1,\ldots m$ iterations
and terminates as soon as there are no more regions to delete. This
is the case, as soon as $MR_{r}^{i}=MR_{r}^{i-1}$ such that $\exists m\geq1\mid\forall r\in R_{color}:\, MR_{r}^{i}=MR_{r}^{i-1}$.

\section{Experimental Results\label{sec:Experimental-Results}}

The proposed algorithm is designed to be parameterless and thus applicable
to different types of images without having to adjust parameter values
by hand. The quality of the resulting segmentation only depends on
the size and resolution of the input image. In this section we discuss
the quality measure for the comparison of different segmentation algorithms
and we show key features, such as the amount of depth of field, that
affects difficulties in segmentation. Further more, we demonstrate
that images with higher resolution generally lead to be better segmentation
results in contrast to the reference algorithms, which loose accuracy
with growing size of the processed image.

In \ref{sub:Internal-parameter} we describe\emph{ }the\emph{ }internal
parameters and threshold values that we determined during our development
and testing phases and show their impact on the quality of the segmentation
result and the performance.

\subsection{Quality measure}

To determine the quality of a segmentation mask \emph{$I$} we use
the \emph{spatial distortion }$d'\left(I,I_{r}\right)$ as proposed
in \cite{kim2005segmenting}:\[
d'\left(I,I_{r}\right)=\frac{\underset{\left(x,y\right)}{\sum}I\left(x,y\right)\otimes I_{r}\left(x,y\right)}{\underset{\left(x,y\right)}{\sum}I_{r}\left(x,y\right)},\]
where $\otimes$ is the binary \emph{XOR} operation and $I_{r}$
is the manually generated reference mask that represents the ground
truth. The spatial distortion denotes the occurred errors, false negatives
and false positives, in relation to the size of the reference mask
$I_{r}$, which is equivalent to the sum of all true positives and
false negatives. Notice that $d'$ can grow larger than $1$ if more
pixels are misclassified than the number of foreground pixels in total.
Let $\textrm{�}$ be a blank mask so that $\textrm{�\ensuremath{\left(x,y\right)}=0}$
for each pixel $p\left(x,y\right)$. The number of false negatives
is now equal to the foreground mask pixel in $I$ because no pixel
in $\textrm{�}$ is defined as forground. For each image $I$, the
spatial distortion $d\left(\textrm{�, I}\right)=1$ . Thus we limit
$d'$ to $d\in\left[0,1\right]$, so that $d\left(I,I_{r}\right)=min\left\{ 1,d'\left(I,I_{r}\right)\right\} .$

\subsection{Dataset}

All experiments were conducted on a diverse dataset of 65 images downloaded
from Flickr and created by our own. The images are from different
categories with strong variations in the amount of depth of field,
as well as in the fuzziness of the background. Also the selection
of the images does not focus on certain sceneries, topics or coloring
schemes in order to avoid overfitting to certain types of images.
In our experiments, we compare the spatial distortion of the proposed
algorithm with re-implementations of the works presented in \cite{kim2005segmenting},
\cite{li2007unsupervized} and \cite{zhang-fuzzy}. The parameters
for all algorithms were optimized to achieve the best average spatial
distortion over the complete test set.

\subsection{Comparison\label{sub:Internal-parameter}}

A major contribution of this algorithm is that none of the parameters
introduced in the previous section needs to be hand tuned for an image
as all parameters are either independent of the image or determined
fully automatically. An overview of the implicit parameters and their
default values can be seen in Table~\ref{tab:Parameters}. %
\begin{table}
\centering{}\caption{Parameters used in the algorithm.\label{tab:Parameters}}
\begin{tabular}{lrr}
\hline 
Parameter & Value & Description\tabularnewline
\hline
$\Theta_{score}$ & 50 & Score $\mu$ threshold of $I_{score}$\tabularnewline
$\Theta_{\epsilon}$ & $\frac{1}{40}$ & Spatial radius of DBSCAN\tabularnewline
$\Theta_{\sigma}$ & $\frac{1}{10}$ & Gaussian blur radius\tabularnewline
$\Theta_{dist}$ & 25 & Color similarity distance\tabularnewline
$\Theta_{rec}$ & $\frac{1}{3}$ & Relative size of reconstruction\tabularnewline
$\Theta_{rel}$ & $\frac{2}{3}$ & Minimum value of relevant regions\tabularnewline
\hline
\end{tabular}
\end{table}

To calculate the score image $I_{score}$ we use a Gaussian blur with
standard deviation of $\Theta_{\sigma}=\frac{1}{10}$. $I_{score}$
is then scaled to fit in $400\times400$ pixel to improve the processing
speed of the subsequent steps without major impact to accuracy. We
set $\Theta_{score}=50$ so that a score $\mu$ must exceed 50 to
be processed by the density-based clustering algorithm DBSCAN that
uses a neighborhood distance $\varepsilon=\Theta_{\epsilon}\sqrt{\left|I\right|}$,
where $\left|I\right|$ is the total pixel count of image \emph{I
}and\emph{ $minPts$} is calculated in dependence of $\varepsilon$,
as described in Sec. \ref{sub:Determination-of-Parameters}. To smooth
the approximate mask we use the morphological operation\emph{ reconstruction
by dilation} $\gamma^{rec}$ with a structuring element \emph{$H$}
of size $\sqrt{\left|I_{approx}\right|}\cdot\Theta_{rec}$ where $\Theta_{rec}=\frac{1}{3}$.
Further we set the maximum distance threshold of similar color values
in the $L^{*}a^{*}b^{*}$ color space to $\Theta_{dist}=25$. The
refinement of the approximate mask removes all regions with a mask
relevance value less than $\Theta_{rel}=\frac{2}{3}$. 

\begin{figure*}
\subfloat[\label{fig:Result-original}Original image.]{\includegraphics[width=0.19\textwidth]{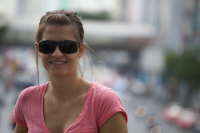}

}\hfill{}\subfloat[Proposed algorithm.]{\includegraphics[width=0.19\textwidth]{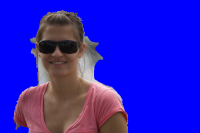}

}\hfill{}\subfloat[Result of fuzzy segmentation \cite{zhang-fuzzy}]{\includegraphics[width=0.19\textwidth]{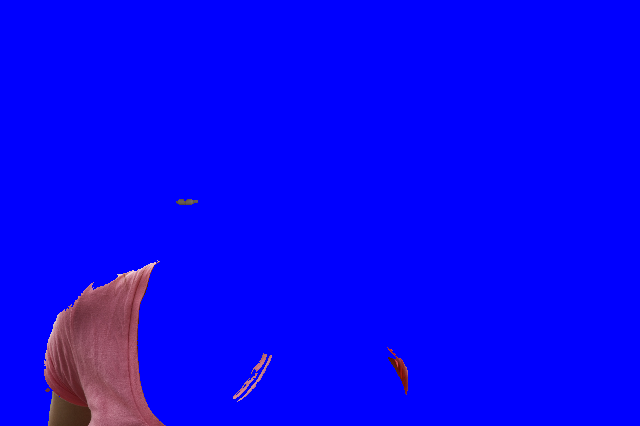}

}\hfill{}\subfloat[Result of morphological segmentation \cite{kim2005segmenting}]{\includegraphics[width=0.19\textwidth]{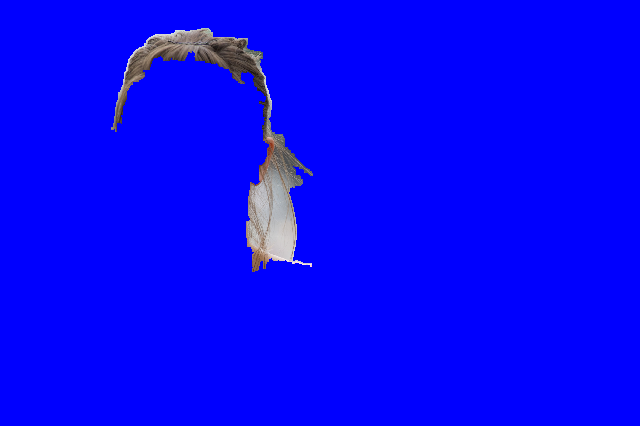}

}\hfill{}\subfloat[\label{fig:Result-video}Result of video segmentation \cite{li2007unsupervized}]{\includegraphics[width=0.19\textwidth]{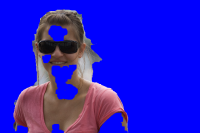}

}\caption{\label{fig:Results}Comparing the results of the different algorithms,
where the input image has complex defocused regions and small DOF.}

\end{figure*}

Fig.~\ref{fig:Results} compares the performance of the reference
algorithms with our proposed method. It can be seen that even though
the computation time of the proposed algorithm is greater than two
of the three reference algorithms, it outperforms the reference algorithms
in terms of spatial distortion in all cases. Also, our algorithm has
an average spatial distortion error of $0.21$ over the complete test
set which is less than half compared to the best competing algorithm
with an average of $0.51$ for the morphological segmentation with
region merging \cite{kim2005segmenting}. Our algorithm also provides
the lowest minimum error of $0.01$ in contrast to $0.02$ for the
fuzzy segmentation algorithm \cite{zhang-fuzzy}.

It should also be noted that in contrast to the reference algorithms,
the proposed algorithm shows improved accuracy with larger images
whereas the competitors loose accuracy with growing size of the image.
The minimum, median, average and standard deviation of the spatial
distortion error are listed in Tab. \ref{tab:dsrVsOthers}.

\begin{figure}
\begin{centering}
\includegraphics[width=0.99\columnwidth]{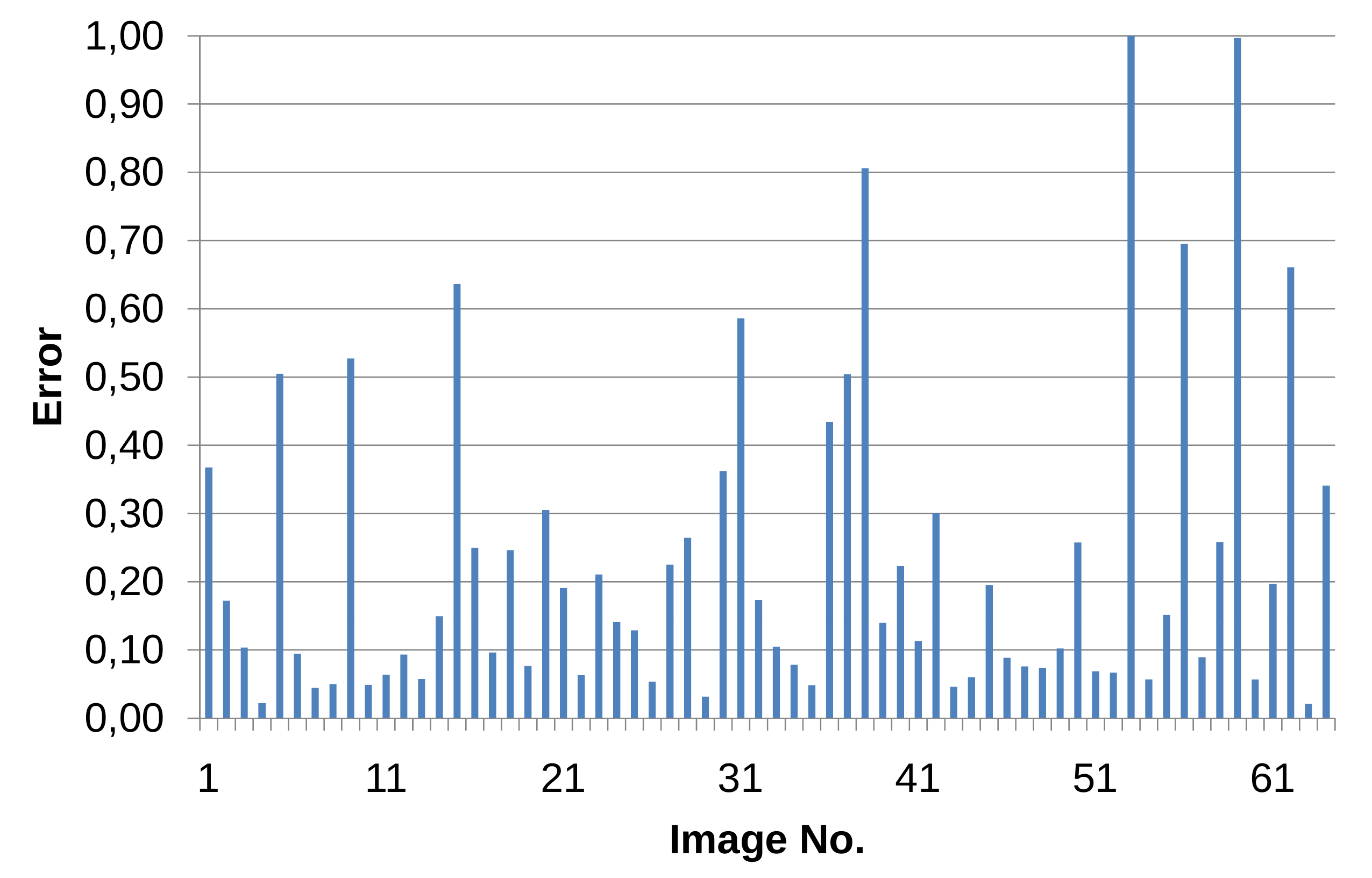}
\par\end{centering}

\caption{\label{fig:dsrVsOthers}Spatial distortion error values of the segmentation
with our proposed algorithm for each of the 65 dataset images.}

\end{figure}

\begin{table}
\centering{}\caption{\label{tab:dsrVsOthers}Spatial distortion and run time of the proposed
algorithm compared to the reference algorithms.}
\begin{tabular}{lrrrr}
\hline 
 & Proposed & \cite{kim2005segmenting} & \cite{zhang-fuzzy} & \cite{li2007unsupervized}\tabularnewline
\hline
Minimum & \textbf{0.01} & 0.05 & 0.02 & 0.07\tabularnewline
Median & \textbf{0.10} & 0.48 & 0.93 & 1\tabularnewline
Average & \textbf{0.21} & 0.51 & 0.84 & 0.81\tabularnewline
Std.Dev. & \textbf{0.21} & 0.26 & 0.25 & 0.31\tabularnewline
Time & 28s & 9s & 54.2s & \textbf{2.7s}\tabularnewline
\end{tabular}
\end{table}
In Fig. \ref{fig:dsrVsOthers} we illustrate all spatial distortion
error values for each segmented image in the dataset.  


\subsection{Image Types\label{sub:Image-Types}}

\begin{figure*}
\subfloat[\label{fig:Result-original-2}Original with homogeneous defocused
regions.]{\includegraphics[width=0.19\textwidth]{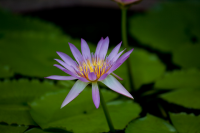}

}\hfill{}\subfloat[Proposed algorithm.]{\includegraphics[width=0.19\textwidth]{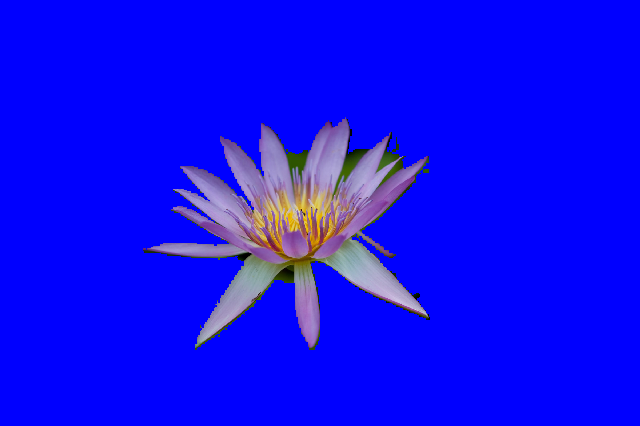}

}\hfill{}\subfloat[Result of fuzzy segmentation \cite{zhang-fuzzy}]{\includegraphics[width=0.19\textwidth]{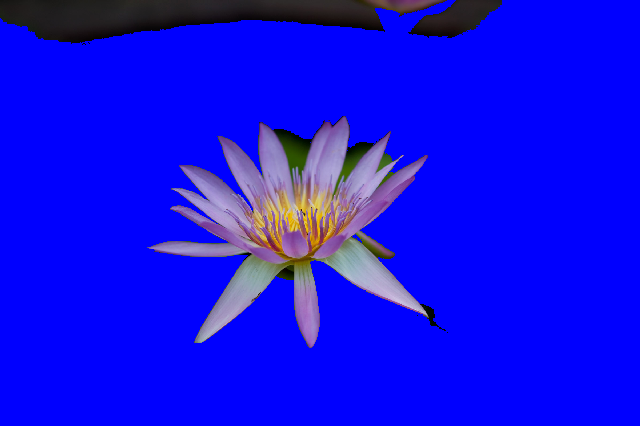}

}\hfill{}\subfloat[Result of morphological segmentation \cite{kim2005segmenting}]{\includegraphics[width=0.19\textwidth]{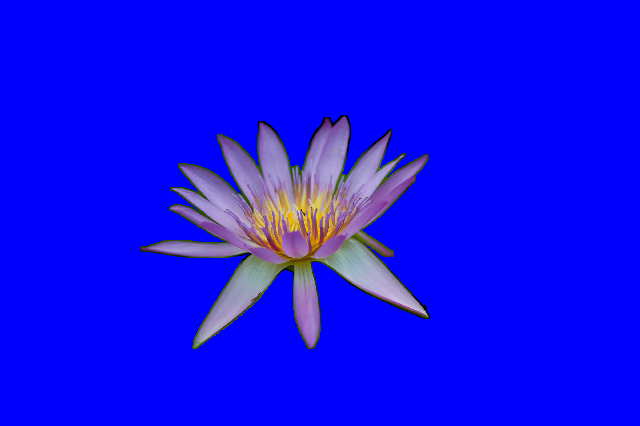}

}\hfill{}\subfloat[\label{fig:Result-video-1}Result of video segmentation \cite{li2007unsupervized}]{\includegraphics[width=0.19\textwidth]{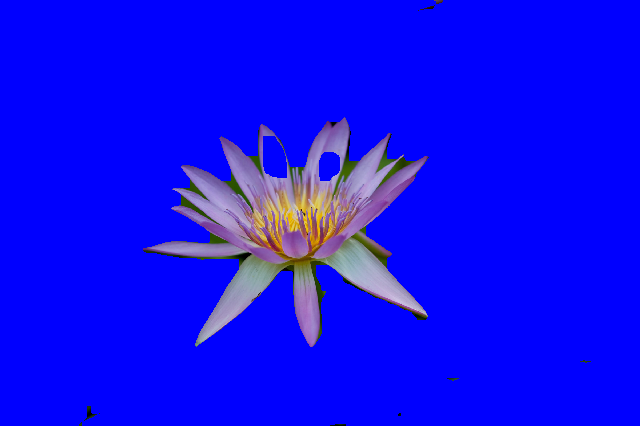}

}\caption{\label{fig:similarResults}Comparing the results of the different
algorithms, where the input image has homogeneous defocused regions,
small DOF and variant colors and thus represents an easy task for
all algorithms.}

\end{figure*}

\begin{figure*}
\subfloat[\label{fig:bird} ]{\includegraphics[width=0.19\textwidth]{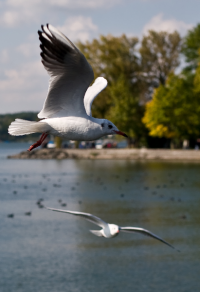}

}\hfill{}\subfloat[\label{fig:our-on-complex}]{\includegraphics[width=0.19\textwidth]{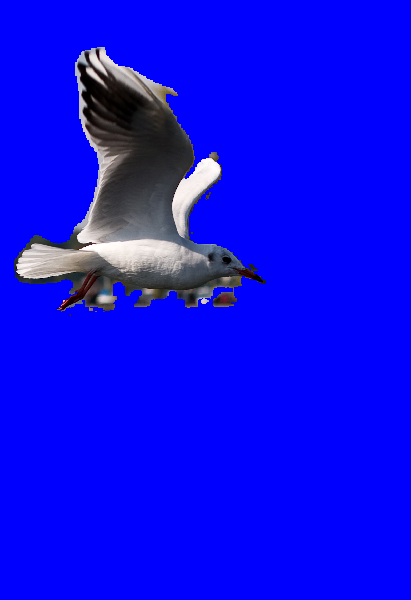}

}\hfill{}\subfloat[\label{fig:fuzzy-on-complex}]{\includegraphics[width=0.19\textwidth]{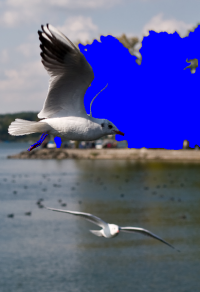}

}\hfill{}\subfloat[\label{fig:morph-on-complex} ]{\includegraphics[width=0.19\textwidth]{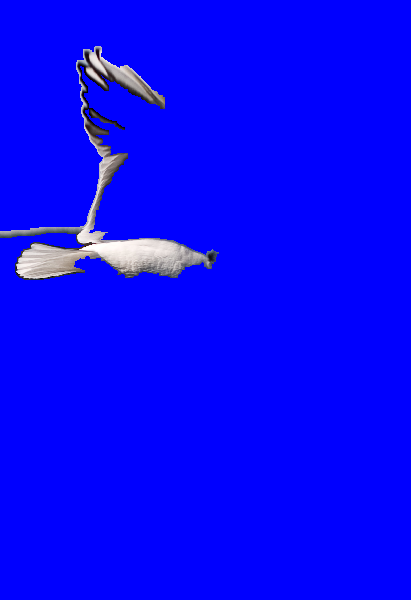}

}\hfill{}\subfloat[\label{fig:video-on-complex}]{\includegraphics[width=0.19\textwidth]{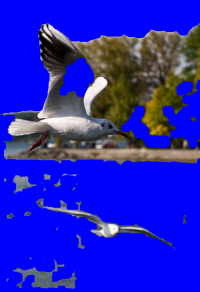}

}\caption{\label{fig:complexBackgroundSegmentation}Results of different segmentation
algorithms applied to an image with complex background (Fig.~.\ref{fig:bird}).
The spatial distortions of the applied algorithms are $0.21$ for
our proposed algorithm (Fig.~\ref{fig:our-on-complex}), $1.0$ by
applying \cite{zhang-fuzzy} (Fig.~\ref{fig:fuzzy-on-complex}),
$0.69$ by applying \cite{kim2005segmenting} (Fig.~\ref{fig:morph-on-complex})
and $1.0$ by applying \cite{li2007unsupervized} (Fig.~\ref{fig:video-on-complex}).}

\end{figure*}

For each image we can define key features as in table \ref{tab:Key-features-of-images}.%
\begin{table}
\centering{}\caption{\label{tab:Key-features-of-images}Key features of an Image}
\begin{tabular}{ccc}
\hline 
amount of DOF & defocused regions & color\tabularnewline
\hline
small & homogeneous & plain\tabularnewline
high & complex & variant\tabularnewline
\hline
\end{tabular}
\end{table}
All these features affect the difficulty for an a accurate segmentation.
Images with smaller DOF, homogeneous defocused regions and variant
colors tend to produce much better segmentation results than images
with high DOF, complex defocused regions and plain colors. The first
type of images is commonly used by most of the competitors' publications
to ensure a high segmentation quality of the presented algorithm.
Figure \ref{fig:similarResults} shows the minor segmentation differences
of a small DOF image.%
A much more challenging task is the segmentation of images with complex
background as shown in Fig.~\ref{fig:complexBackgroundSegmentation}.
Thereby our proposed algorithm achieves a good segmentation result
with a spatial distortion of $0.21$ (c.f. Fig.~\ref{fig:our-on-complex}),
where the other segmentation algorithms \cite{zhang-fuzzy}, \cite{kim2005segmenting}
and \cite{li2007unsupervized} fail with spatial distortion values
of each larger than $0.69$ (c.f. Fig.~\ref{fig:fuzzy-on-complex},
\ref{fig:morph-on-complex} and \ref{fig:video-on-complex}).

\subsection{Size of input image\label{sub:Size-of-imput}}

\begin{figure}
\begin{centering}
\includegraphics[width=0.99\columnwidth]{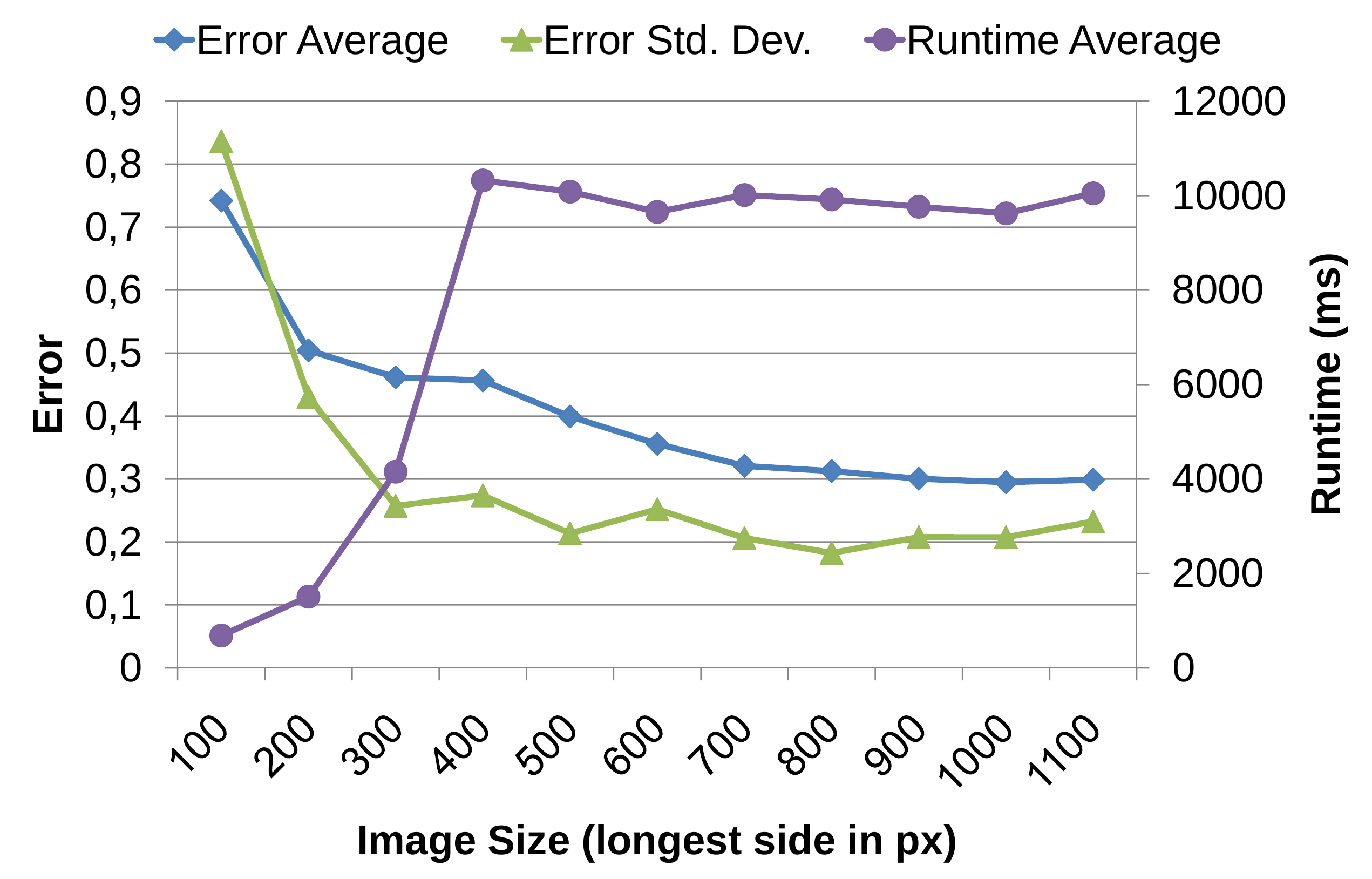}
\par\end{centering}

\caption{\label{fig:sizecomparison}Impact of the size of the input image to
the error rate of the segmentation measured by the spatial distortion
and the average runtime per image.}

\end{figure}

One of the most influential variables on segmentation quality is the
resolution of the input image. A comparatively high resolution is
needed for a proper segmentation, if for example an image has just
a slightly defocused background and thus shows significant texture.
Thus we designed the algorithm to be able to handle a large scope
of resolutions properly without loss of quality. By using the image
of Fig.~\ref{fig:bird} as input, a spatial distortion value of $0.65$
can already be achieved at the relatively small resolution of $200\times300$
pixels. For this particular image, a resolution of $240\times350$
is needed to lower the spatial distortion to $0.42$. Fig.~\ref{fig:sizecomparison}
shows the diversification of average and standard deviation of the
spatial distortion depending on the increase of image size. Because
the orientation (portrait or landscape) and aspect ratio (2:3, 4:3,
etc.) of the images in our test set varies, we rescale each image
so that its longest side equals the value of the resizing operation.
In the following we denote the term image size by the longer side
of the image.

As we increase the input image size from 100 pixels to 200 pixels,
we can lower the average spatial distortion from $0.74$ to $0.5$
and the standard deviation of the spatial distortion from $0.84$
to $0.43$. As the size of the images reach 600 pixels, the average
and standard deviation of the spatial distortion are $0.36$ and $0.25$,
respectively. At the same time, the runtime increases from an average
of 686 milliseconds per image at 100 pixels, to 4155 milliseconds
at 300 pixels. In our experiments we could also show a significant
decrease of the average spatial distortion and the corresponding standard
deviation up to an image size of about $600$ pixels. For images larger
than $800$ pixels, the average spatial distortion and standard deviation
still improve, yet at a significantly slower rate than in the case
of smaller images. In Fig.~\ref{fig:sizecomparison} we summarize
the results of this experiment. Because the score image $I_{score}$
was scaled to fit in $400\times400$ pixels in the score clustering
stage, the exponential runtime of the algorithm stagnates after reaching
image sizes $\geq400$ pixels for its longest side.

\subsection{Sample Segmentations}

\begin{figure*}
\includegraphics[width=0.19\textwidth]{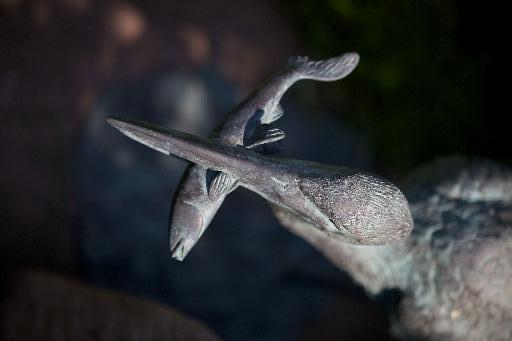}\hfill{}\includegraphics[width=0.19\textwidth]{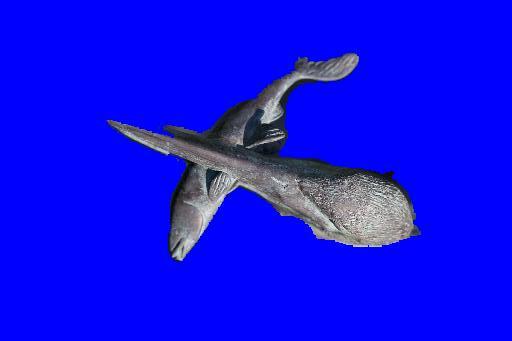}\hfill{}\includegraphics[width=0.19\textwidth]{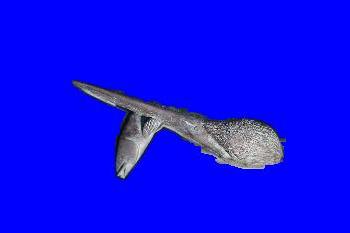}\hfill{}\includegraphics[width=0.19\textwidth]{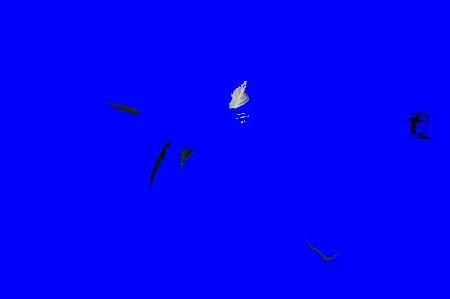}\hfill{}\includegraphics[width=0.19\textwidth]{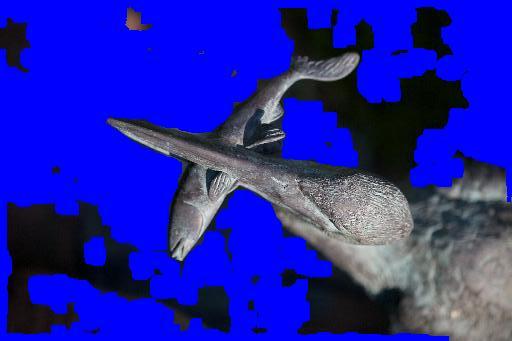}\hfill{}\medskip{}
\includegraphics[width=0.19\textwidth]{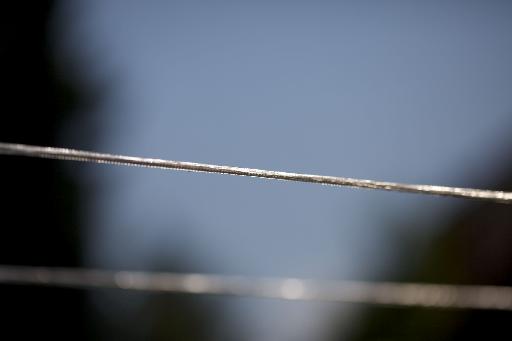}\hfill{}\includegraphics[width=0.19\textwidth]{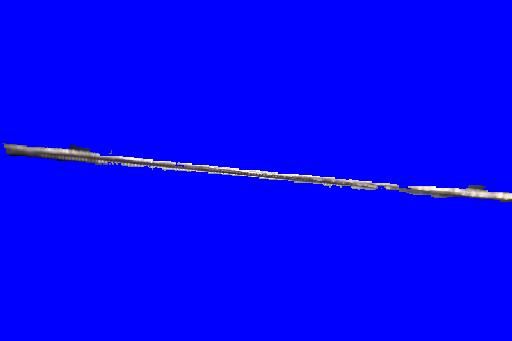}\hfill{}\includegraphics[width=0.19\textwidth]{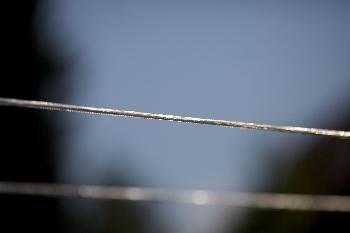}\hfill{}\includegraphics[width=0.19\textwidth]{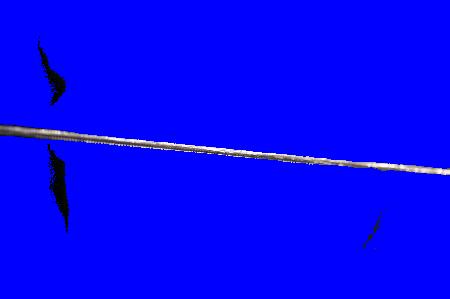}\hfill{}\includegraphics[width=0.19\textwidth]{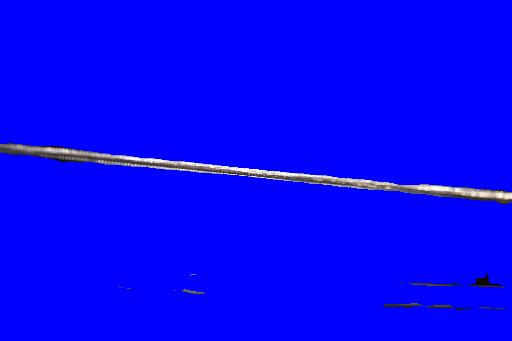}\hfill{}\medskip{}
\includegraphics[width=0.19\textwidth]{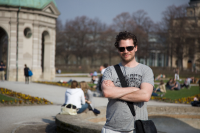}\hfill{}\includegraphics[width=0.19\textwidth]{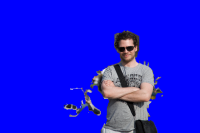}\hfill{}\includegraphics[width=0.19\textwidth]{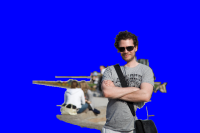}\hfill{}\includegraphics[width=0.19\textwidth]{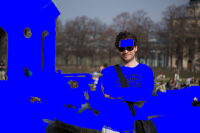}\hfill{}\includegraphics[width=0.19\textwidth]{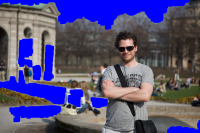}\hfill{}\medskip{}
\includegraphics[width=0.19\textwidth]{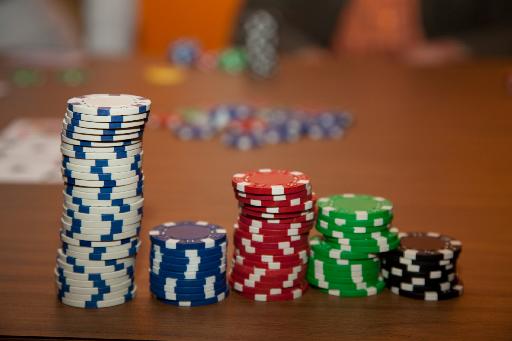}\hfill{}\includegraphics[width=0.19\textwidth]{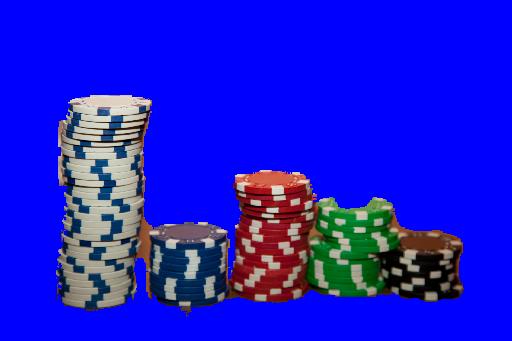}\hfill{}\includegraphics[width=0.19\textwidth]{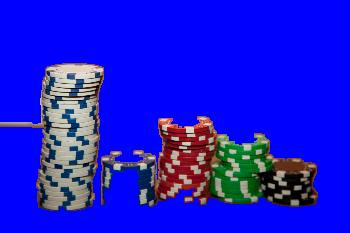}\hfill{}\includegraphics[width=0.19\textwidth]{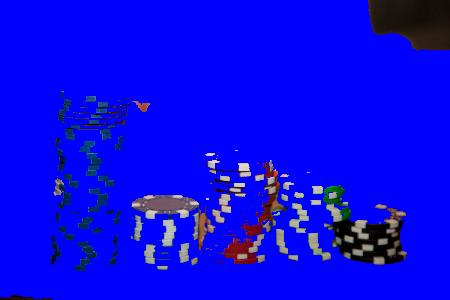}\hfill{}\includegraphics[width=0.19\textwidth]{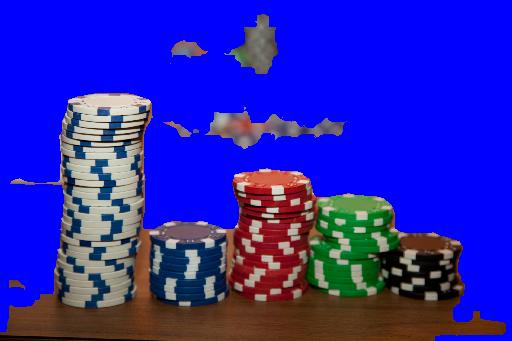}\hfill{}\medskip{}
\includegraphics[width=0.19\textwidth]{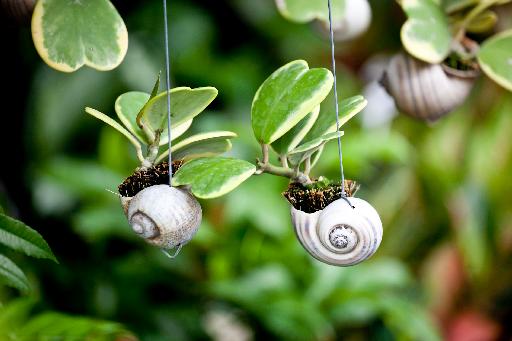}\hfill{}\includegraphics[width=0.19\textwidth]{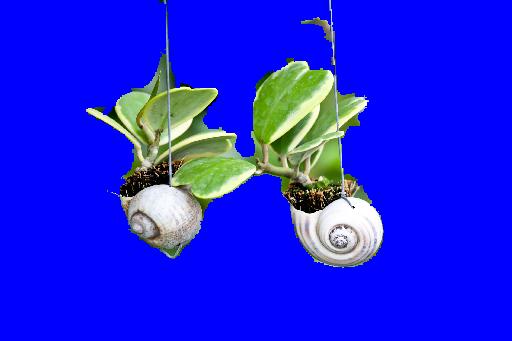}\hfill{}\includegraphics[width=0.19\textwidth]{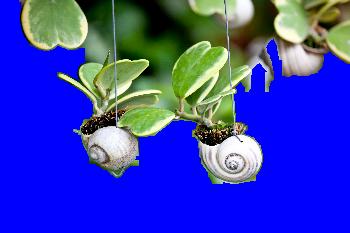}\hfill{}\includegraphics[width=0.19\textwidth]{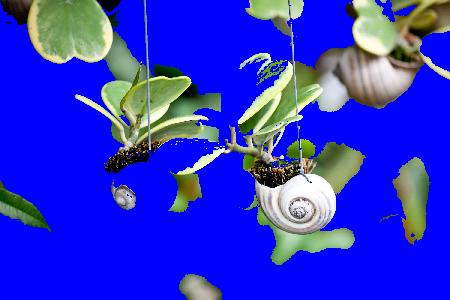}\hfill{}\includegraphics[width=0.19\textwidth]{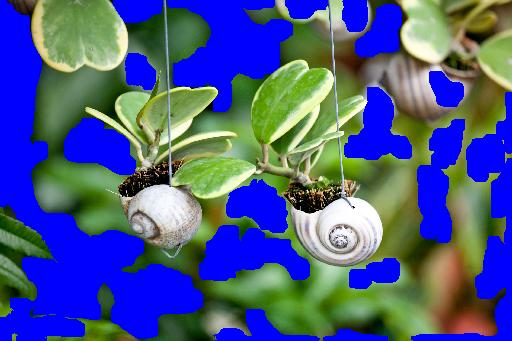}\hfill{}\medskip{}
\includegraphics[width=0.19\textwidth]{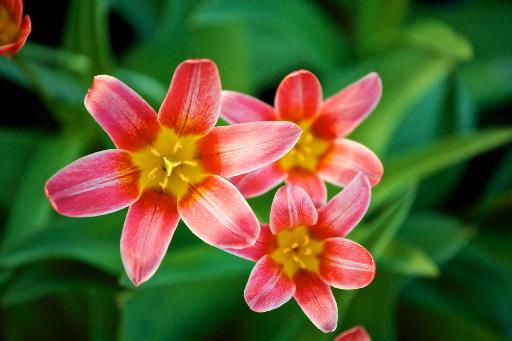}\hfill{}\includegraphics[width=0.19\textwidth]{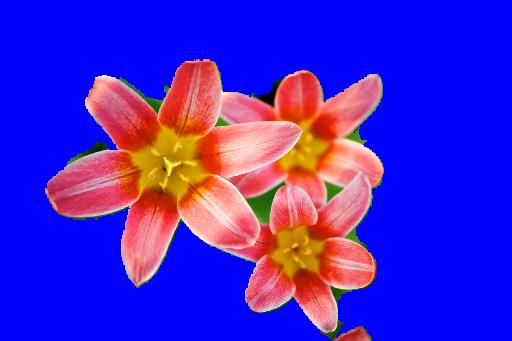}\hfill{}\includegraphics[width=0.19\textwidth]{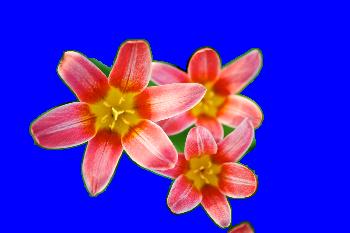}\hfill{}\includegraphics[width=0.19\textwidth]{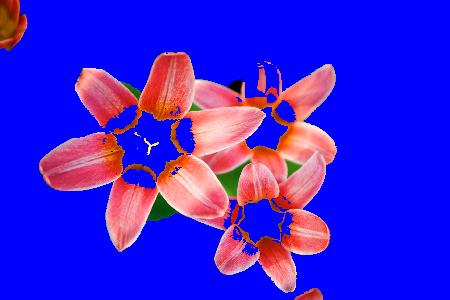}\hfill{}\includegraphics[width=0.19\textwidth]{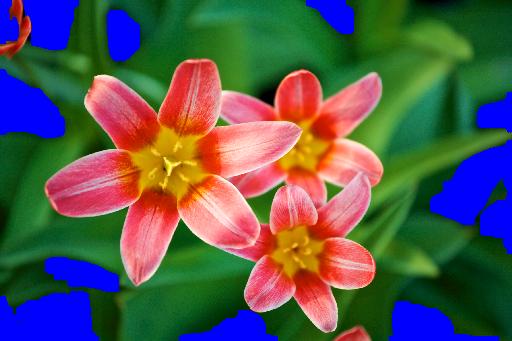}\hfill{}\medskip{}
\includegraphics[width=0.19\textwidth]{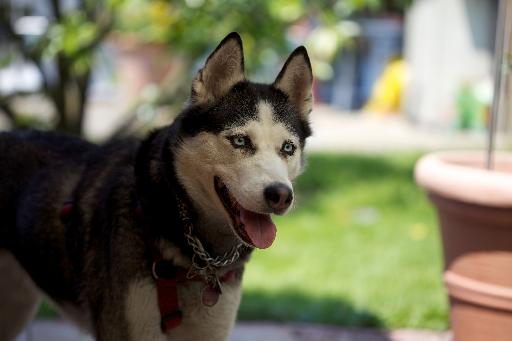}\hfill{}\includegraphics[width=0.19\textwidth]{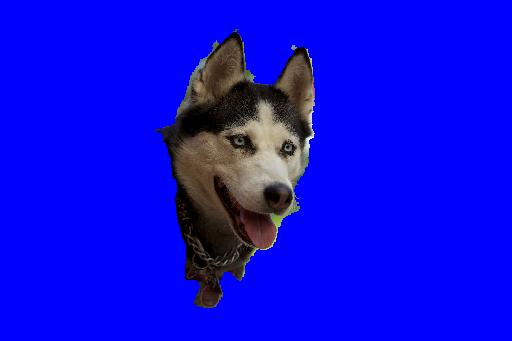}\hfill{}\includegraphics[width=0.19\textwidth]{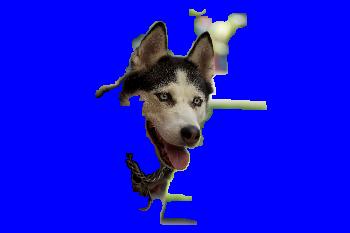}\hfill{}\includegraphics[width=0.19\textwidth]{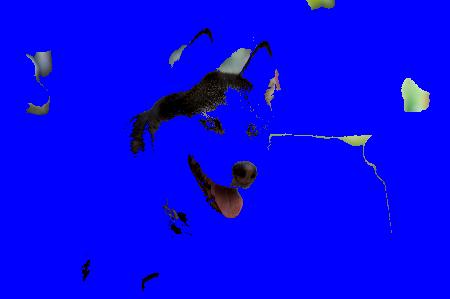}\hfill{}\includegraphics[width=0.19\textwidth]{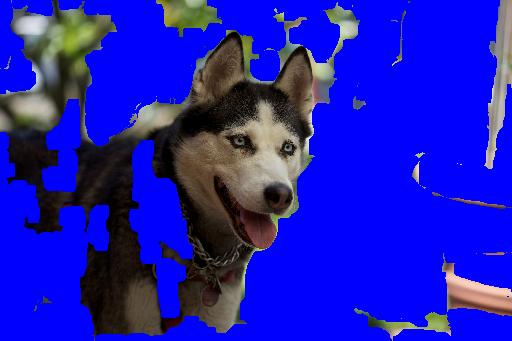}\hfill{}\medskip{}
\includegraphics[width=0.19\textwidth]{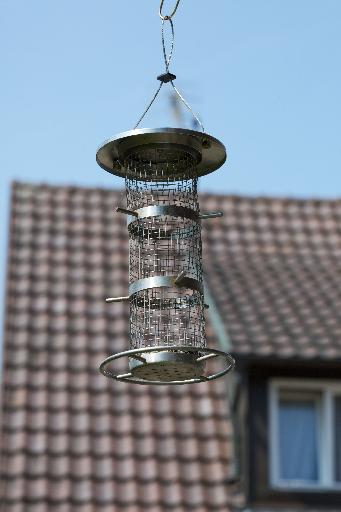}\hfill{}\includegraphics[width=0.19\textwidth]{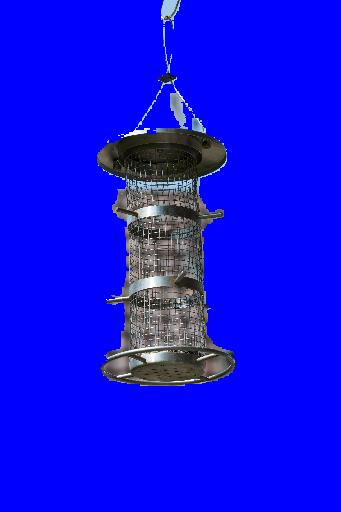}\hfill{}\includegraphics[width=0.19\textwidth]{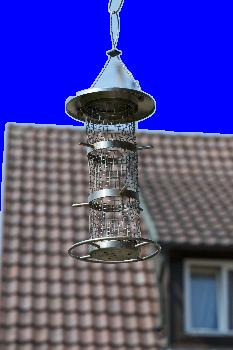}\hfill{}\includegraphics[width=0.19\textwidth]{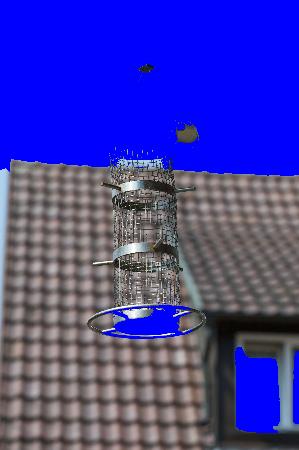}\hfill{}\includegraphics[width=0.19\textwidth]{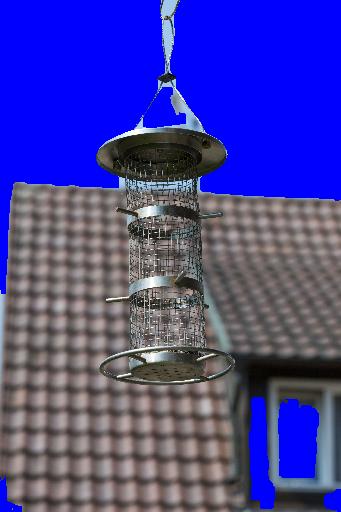}\hfill{}

\centering{}\caption{\label{fig:versus}Segmentation results of different color images
(first column) by applying our proposed algorithm (second column),
\cite{kim2005segmenting} (third column), \cite{zhang-fuzzy} (fourth
column) and \cite{li2007unsupervized} (fifth column).}

\end{figure*}

Fig. \ref{fig:versus} presents some segmentation results of different
color images. The input image is shown in the first column. In the
second column our segmentation result is depicted. The morphological
segmentation \cite{kim2005segmenting} is placed in the third column,
fuzzy segmentation \cite{zhang-fuzzy} in the fourth column and the
video segmentation \cite{li2007unsupervized} in the last column.
Notice, that in one case the resulting mask covers the entire image
(as seen in the second row of Fig.~\ref{fig:versus} in the third
column).

\section{Parameter Settings\label{sec:Parameter-Settings}}

In this section, we describe\emph{ }the\emph{ }internal\emph{ }parameter
settings and threshold values. The choice of the parameter values
is the result of an extensive evaluation and represents a recommended
set of values that produced the best and stable results. Thus, none
of these values has to be set by the user. For each parameter we discuss
the impact on the average and standard deviation of the spatial distortion
error and on the performance by evaluating the segmentation on our
data set.

\subsection{Blur Radius}

\begin{figure}
\begin{centering}
\includegraphics[width=0.99\columnwidth]{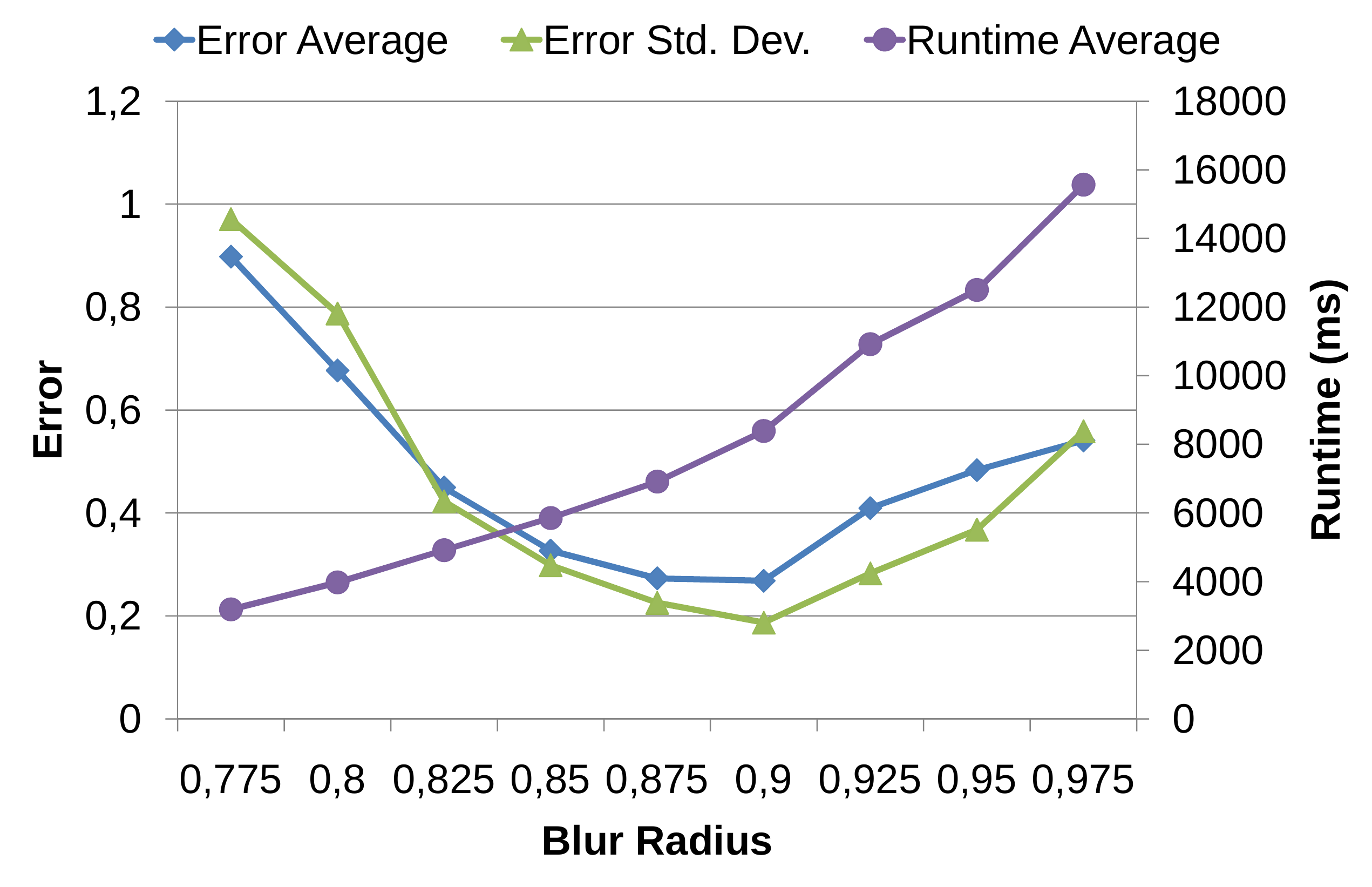}
\par\end{centering}

\caption{\label{fig:sigmaBlur}Impact of the blur radius $\Theta_{\sigma}$
on the spatial distortion error and average runtime per image.}

\end{figure}

In the first stage, called \emph{deviation scoring}, the parameter
$\Theta_{\sigma}$ determines the size of the radius of the Gaussian
convolution kernel to remove noise from the input image. The same
kernel is used to re-blur the de-noised image to compute the deviation
of the mean neighbor difference of each pixel in the de-noised and
re-blurred Image. Fig.~\ref{fig:sigmaBlur} shows the distribution
of the spatial distortion on the primary y-axis with the mean execution
time of the algorithm in milliseconds being displayed on the secondary
y-axis. If $\Theta_{\sigma}$ is set too low, too few noise is removed
from the input image. In addition, edges of the OOI in the re-blurred
image have a reduced occurrence and thus can be determined less effective
in contrast to the edges of the background noise. As illustrated in
Fig.~\ref{fig:sigmaBlur}, a small blur radius of $\Theta_{\sigma}=0.775$
results in a high average spatial distortion $d>0.8$ compared to
results of a segmentation with a more convenient choice of $\Theta_{\sigma}$.
We assign $\Theta_{\sigma}=0.9$ as the best trade-off between runtime
(approximately 8s per image) and an average spatial distortion of
$d_{avg}=0.26$. Larger values of $\Theta_{\sigma}$ cause a very
intense smoothing operation, so that the edges of the OOI loose their
significance.%

\subsection{Score Clustering Threshold}

The score clustering threshold $\Theta_{score}$ describes the minimum
score value $\mu$, that each pixel has to exceed in order to be processed
by the DBSCAN clustering. This parameter ensures that the clustering
algorithm does not have to handle a too large number of pixels with
a score value which is not significant. If $\Theta_{score}$ is set
too low, DBSCAN will consume more time without improving the segmentation
quality significantly. In cases of $\Theta_{score}<25$ the spatial
distortion grows even higher because too many pixels with insignificant
score values are considered.

Otherwise if $\Theta_{score}$ is set too high, too many pixels with
a potentially essential score value are not processed and the remaining
clusters contain too few points. This results in a smaller mask that
does not cover the whole focus area.%
\begin{figure}
\begin{centering}
\includegraphics[width=0.99\columnwidth]{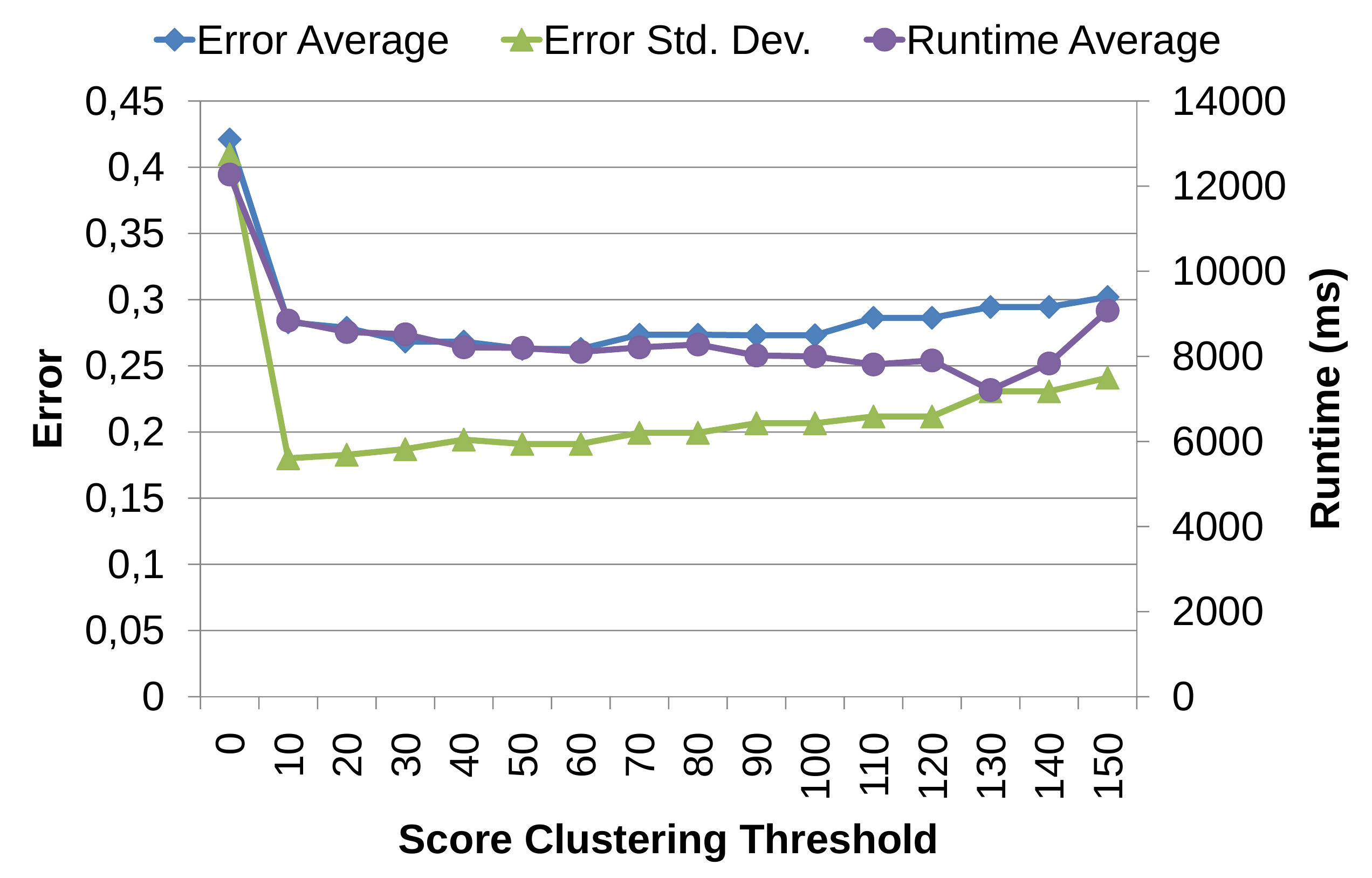}
\par\end{centering}

\caption{\label{fig:tethaScore}Impact of the score clustering threshold $\Theta_{score}$
on the spatial distortion error and the average runtime per image.}

\end{figure}
As can be seen in Fig.~\ref{fig:tethaScore}, $\Theta_{score}=50$
can be defined as an optimal trade-off between spatial distortion
and runtime.

\subsection{Neighborhood distance}

\begin{figure*}
\subfloat[\label{fig:bird-1}$I_{score}$ from Lena image]{\includegraphics[width=0.23\textwidth]{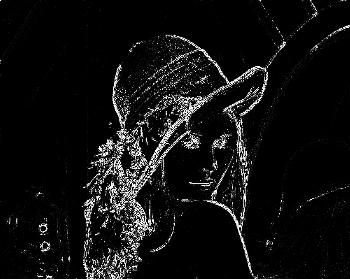}

}\hfill{}\subfloat[\label{fig:thetaEpsilon_low}Clustering result with \foreignlanguage{american}{$\Theta_{\varepsilon}=0.005$} ]{\includegraphics[width=0.23\textwidth]{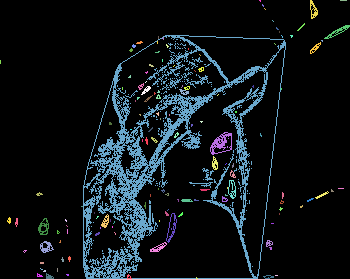}

}\hfill{}\subfloat[\label{fig:thetaEpsilon_mean}Clustering result with \foreignlanguage{american}{$\Theta_{\varepsilon}=0.025$}
.]{\includegraphics[width=0.23\textwidth]{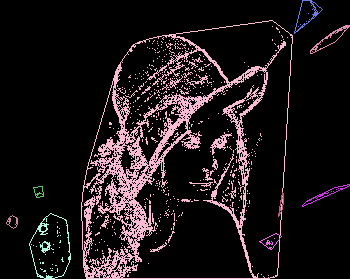}

}\hfill{}\subfloat[\label{fig:thetaEpsilon_high}Clustering result with \foreignlanguage{american}{$\Theta_{\varepsilon}=0.045$} ]{\includegraphics[width=0.23\textwidth]{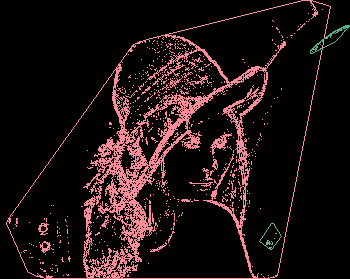}

}\caption{\label{fig:lena}Impact of the \foreignlanguage{american}{$\Theta_{\varepsilon}$}
parameter on score clustering. For better visual distinction, each
cluster is represented in a random color and surrounded by its convex
hull.}

\end{figure*}

The neighborhood distance $\Theta_{\varepsilon}$ directly influences
the size of the $\varepsilon$ parameter of DBSCAN, as $\varepsilon=\sqrt{\left|I\right|}\cdot\Theta_{\varepsilon}$.
A higher value of $\Theta_{\varepsilon}$ increases the spatial radius
$\varepsilon$ so that a core point has a larger reachability distance.
If $minPts$, DBSCAN's second parameter, remains unchanged, an increase
in $\Theta_{\varepsilon}$ would result in a decreasing number of
larger clusters. As \emph{$minPts$} is defined relatively to $\varepsilon$
(see Sec. \ref{sub:Determination-of-Parameters}) an increase of $\Theta_{\varepsilon}$
would also enlarge \emph{$minPts$ }and vice versa. Fig.~\ref{fig:lena}
illustrates the different clustering results when changing this parameter.
If $\Theta_{\varepsilon}$ is too low, the main focus area is split
into many different, mostly small clusters. Thus, important information
in $I_{score}$ would be interpreted as noise, as shown in Fig.~\ref{fig:thetaEpsilon_low}.

\begin{figure}
\begin{centering}
\includegraphics[width=0.99\columnwidth]{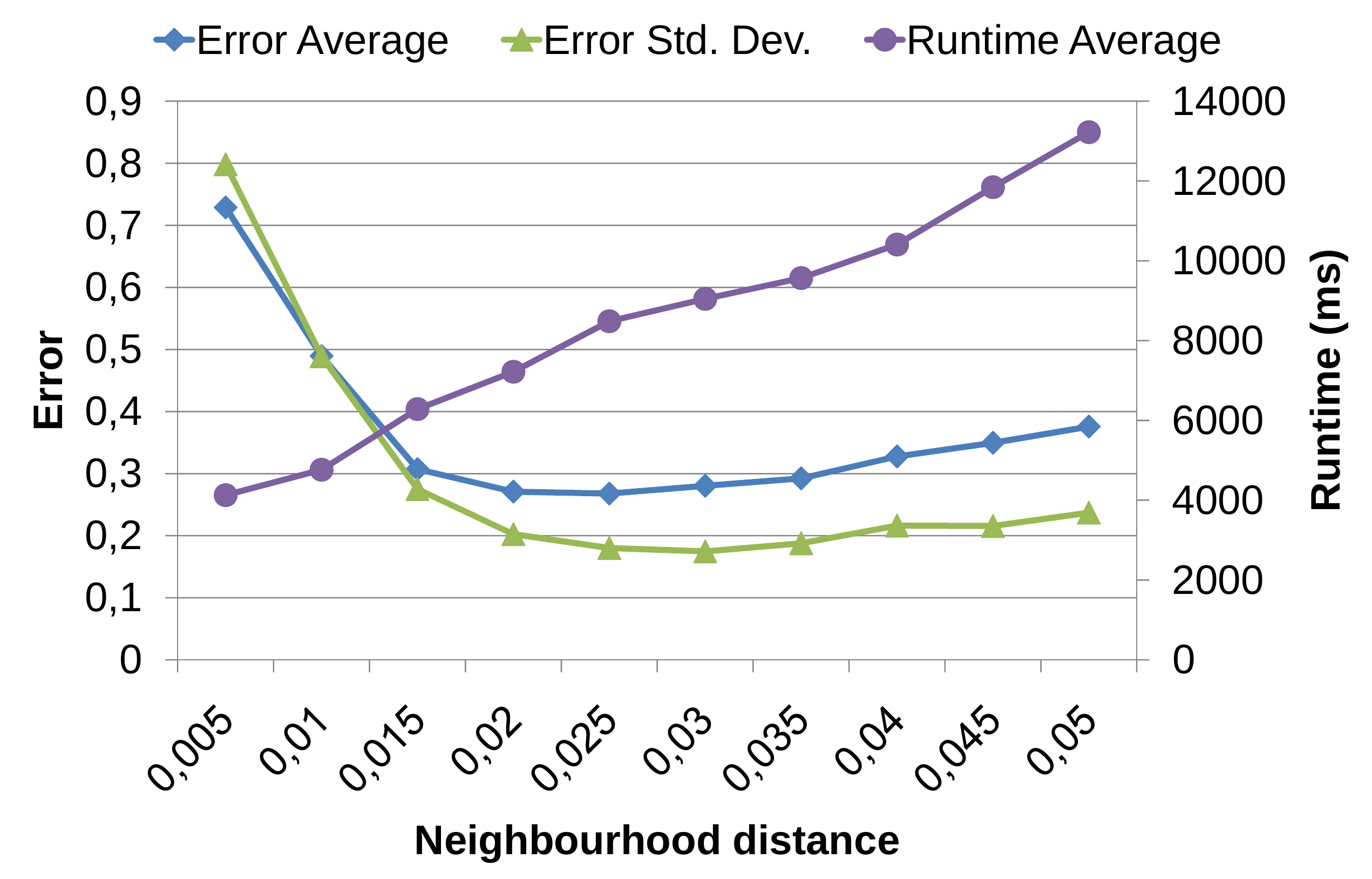}
\par\end{centering}

\caption{\label{fig:tethaEpsilon}Impact of the neighborhood distance $\Theta_{\varepsilon}$
on the spatial distortion error and the average runtime per image.}

\end{figure}

As you can see in \ref{fig:thetaEpsilon_mean}, $\Theta_{\varepsilon}=0.025$
produced a much more reliable clustering result and covers the main
focus region by not including surrounding noise. If $\Theta_{\varepsilon}$
is set too high, too much noise surrounding the OOI is merged with
the main cluster (see \ref{fig:thetaEpsilon_high}). The influence
of $\Theta_{\varepsilon}$ on the segmentation result is shown in
figure \ref{fig:tethaEpsilon}.%
As optimum distance we set $\Theta_{\varepsilon}=0.025$, where the
average and standard deviation of the spatial distortion are $0.26$
and $0.18$, and the average runtime$<9000$ milliseconds per image
is still acceptable.

\subsection{Size of the structuring element}

\begin{figure*}
\subfloat[\label{fig:sport_org}]{\includegraphics[width=0.23\textwidth]{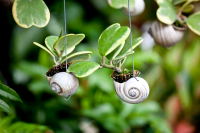}

}\hfill{}\subfloat[\label{fig:sport_smallClose}]{\includegraphics[width=0.23\textwidth]{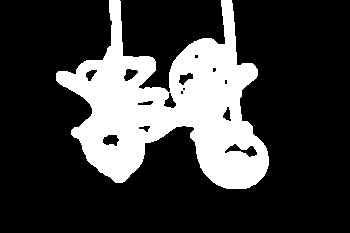}

}\hfill{}\subfloat[\label{fig:sport_mediumClose}]{\includegraphics[width=0.23\textwidth]{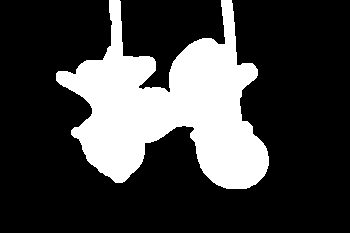}

}\hfill{}\subfloat[\label{fig:sport_highClose}]{\includegraphics[width=0.23\textwidth]{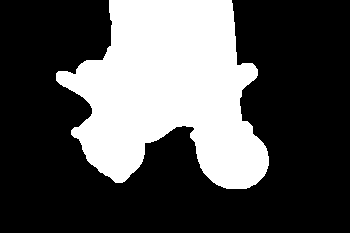}

}\caption{\label{fig:structurinElementParameter}Impact of the parameter $\Theta_{rec}$
on smoothing and filling the approximate mask of the input image (Fig.~\ref{fig:sport_org}).
Fig. \ref{fig:sport_smallClose}-\ref{fig:sport_highClose} show the
approximate masks after \emph{reconstruction by dilation} with small
($\Theta_{rec}=0.1$), medium ($\Theta_{rec}=0.25$) and large ($\Theta_{rec}=0.6$)
structuring elements.}

\end{figure*}

\begin{figure}
\begin{centering}
\includegraphics[width=0.99\columnwidth]{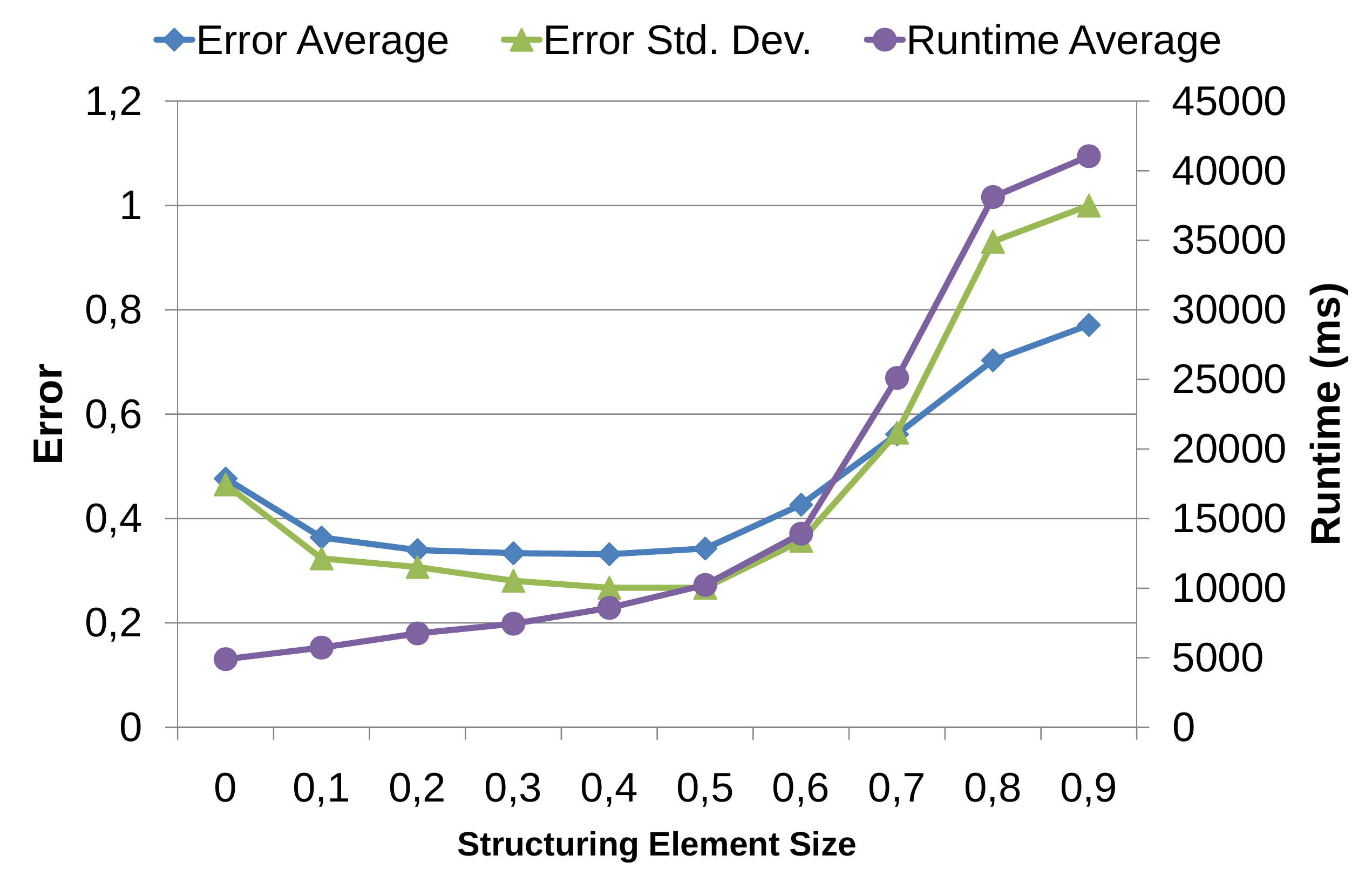}
\par\end{centering}

\caption{\label{fig:tethaRec}Impact of the neighborhood distance $\Theta_{rec}$
to the spatial distortion error and the average runtime per image.}

\end{figure}

The size of the structuring element \emph{$H$} is used in our \emph{mask
approximation} stage after the \emph{convex hull linking} step by
morphological filter operations, in order to smooth the current mask.
If \emph{$H$} is too small, little gaps remain and prevent the following
\emph{reconstruction by dilation} operation $\gamma^{rec}$ from filling
holes in the approximate mask as only dark regions that are completely
surrounded by the white mask are treated as holes. As the size of
\emph{$H$} increases, more gaps are closed and the contour of the
OOIs approximate mask gets more fuzzy. The operation\emph{ reconstruction
by dilation} $\gamma^{rec}$ uses a structuring element \emph{$H'$}.
The size of the structuring element is calculated by $\sqrt{\left|I\right|}\cdot\Theta_{rec}$.
In summary it can be said that smaller sizes of $H'$ only cause the
filling of small holes in the mask, while larger sizes of $H'$ fill
larger holes also. Fig.~\ref{fig:structurinElementParameter} shows
the operation $\gamma^{rec}$ with small, medium and large structuring
elements.%
For $\Theta_{rec}\in\left[\frac{1}{5},\frac{1}{2}\right]$ the overall
segmentation result of the complete data set is not highly influenced
by $\Theta_{rec}$ as shown in Fig.~\ref{fig:tethaRec}. Values outside
of that range produce a higher error rate. As $\Theta_{rec}$ approaches
$1$, the average runtime also increases rapidly from around $\approx10$
seconds at $\Theta_{rec}=\frac{1}{2}$ to $>40$ seconds at $\Theta_{rec}=\frac{9}{10}$
.

\subsection{Color similarity distance}

\begin{figure}
\begin{centering}
\includegraphics[width=0.99\columnwidth]{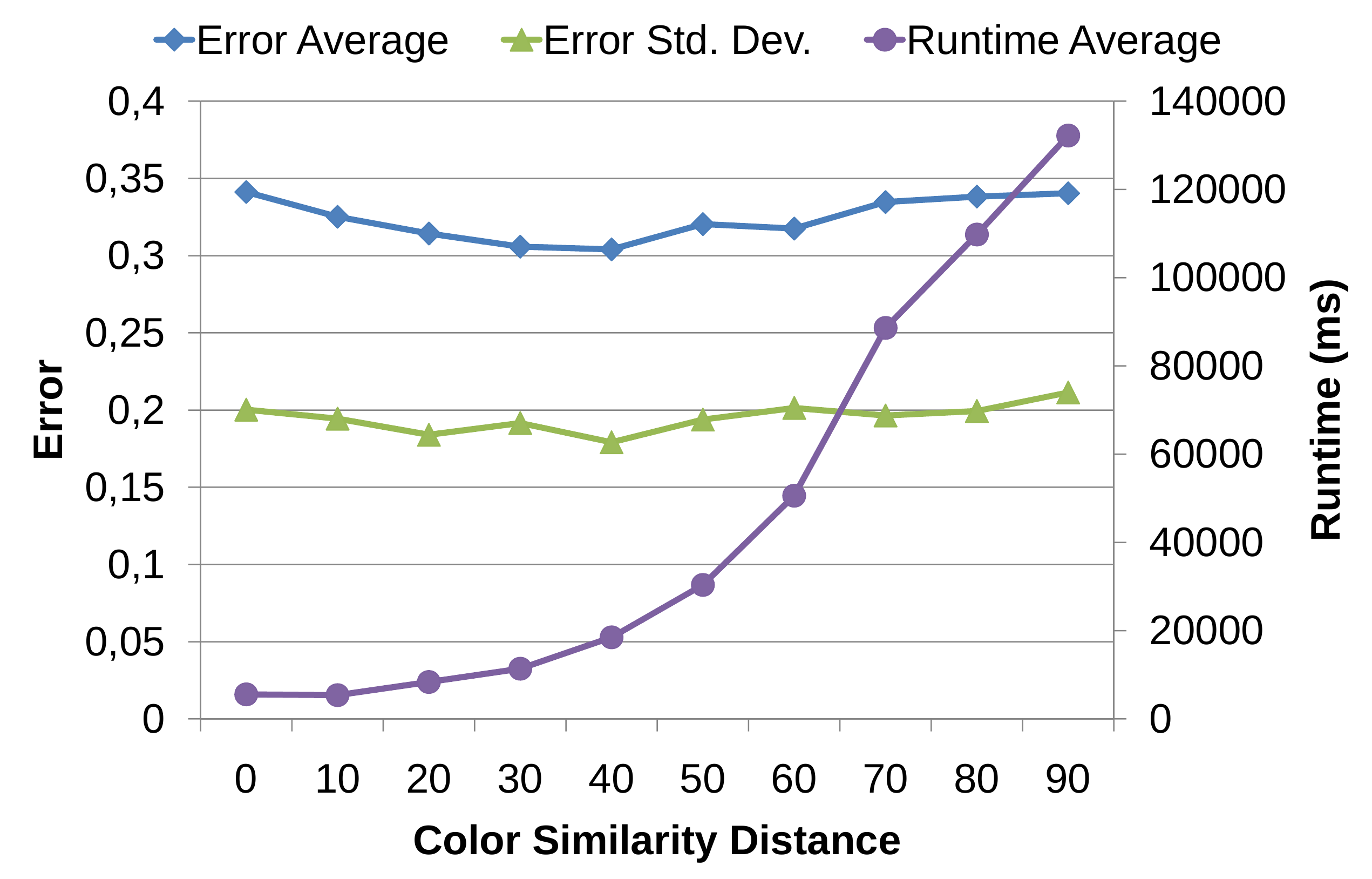}
\par\end{centering}

\caption{\label{fig:tethaDist}Impact of the color similarity distance $\Theta_{dist}$
on the spatial distortion error and the average runtime per image.}

\end{figure}

The color similarity parameter $\Theta_{dist}$ describes the distance
$\Delta E^{*}\left(u,v\right)$ that two colors $u,v$ of the $L^{*}a^{*}b^{*}$
color space may not exceed in order to be considered as similar. Thus,
the amount of $\Theta_{dist}$ has direct impact on the amount of
color regions which are extracted from the original images. If $\Theta_{dist}$
is set to a very low value, the resulting color regions are very small.
This causes large relative \emph{mask relevance }values in case of
small color regions which are overlapped by a small area of the approximation
mask. And thus, less regions are removed in the following \emph{region
scoring} stage. Too large values for $\Theta_{dist}$ imply that
too many regions are merged together and thus the \emph{region scoring
}will become too vague.%
\begin{figure*}
\subfloat[\label{fig:colordistparameter_input}]{\includegraphics[width=0.23\textwidth]{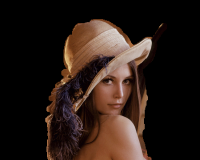}

}\hfill{}\subfloat[\label{fig:colordistparameter_small}]{\includegraphics[width=0.23\textwidth]{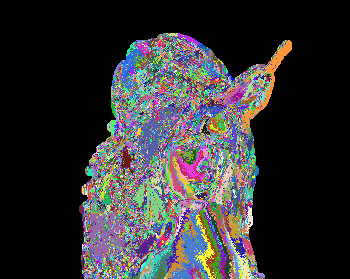}

}\hfill{}\subfloat[\label{fig:colordistparameter_medium}]{\includegraphics[width=0.23\textwidth]{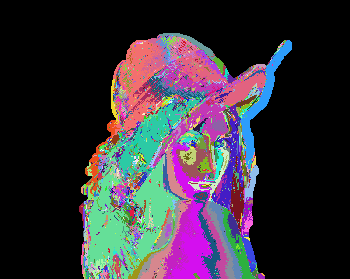}

}\hfill{}\subfloat[\label{fig:colordistparameter_large}]{\includegraphics[width=0.23\textwidth]{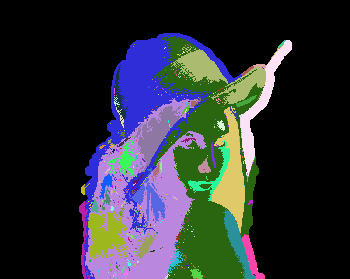}

}\caption{\label{fig:colorDistanceParameter}Impact of the parameter $\Theta_{dist}$
on grouping the smoothed approximation mask into regions of similar
colors of the input image (Fig.~\ref{fig:colordistparameter_input}).
Fig.~\ref{fig:colordistparameter_small}-Fig.~\ref{fig:colordistparameter_large}
show the approximate mask with small ($\Theta_{dist}=5$), medium
($\Theta_{dist}=25$) and large ($\Theta_{dist}$=40) values. For
better visual distinction, each region is represented in a random
color.}

\end{figure*}
Fig.~\ref{fig:colorDistanceParameter} shows the different results
of the disjoint color regions by altering the parameter $\Theta_{dist}$.
Fig.~\ref{fig:colordistparameter_input} is the pixel subset of the
original image marked by the smoothed mask which is returned by the
\emph{mask approximation }stage (Sec.~\ref{sub:Mask-Approximation})
of the algorithm. If $\Theta_{dist}$ is set to a small value as in
Fig.~\ref{fig:colordistparameter_small}, a large amount of disjoint
color regions is created compared to Fig.~\ref{fig:colordistparameter_medium}
where $\Theta_{dist}=25$. Very few color regions are constructed
if a large value like $\Theta_{dist}=40$ is applied to the \emph{color
segmentation }stage. This is illustrated in Fig.~\ref{fig:colordistparameter_large}.
Usually, this makes the \emph{region scoring }stage (Sec.~\ref{sub:Region-Scoring})
more vague. The reason for this is that in the case of a large $\Theta_{dist}$
it is more likely that $n>1$ regions $r_{1},\ldots,r_{n}$ with high
score variation \[
min\left\{ MR_{r_{1}},\ldots,MR_{r_{n}}\right\} \ll max\left\{ MR_{r_{1}},\ldots,MR_{r_{n}}\right\} \]
are merged together in one larger region $r=r_{1}\cup\ldots\cup r_{n}$,
where $MR_{r_{i}},\, i=1\ldots n$ denotes the \emph{mask relevance
}introduced in Sec.~\ref{sub:Region-Scoring}. The deletion\emph{
}of\emph{ $r$} would then generate more false negatives, which means
that the resulting final mask does not cover the entire OOI. The inclusion
of \emph{$r$ }would generate more false positives, so that more background
is finally included. As illustrated in Fig.~\ref{fig:tethaDist},
an increasing value of $\Theta_{dist}$ also causes an increased processing
time. For our tests, we chose $\Theta_{dist}=25$.

\subsection{Region Scoring Relevance }

\begin{figure}
\begin{centering}
\includegraphics[width=0.99\columnwidth]{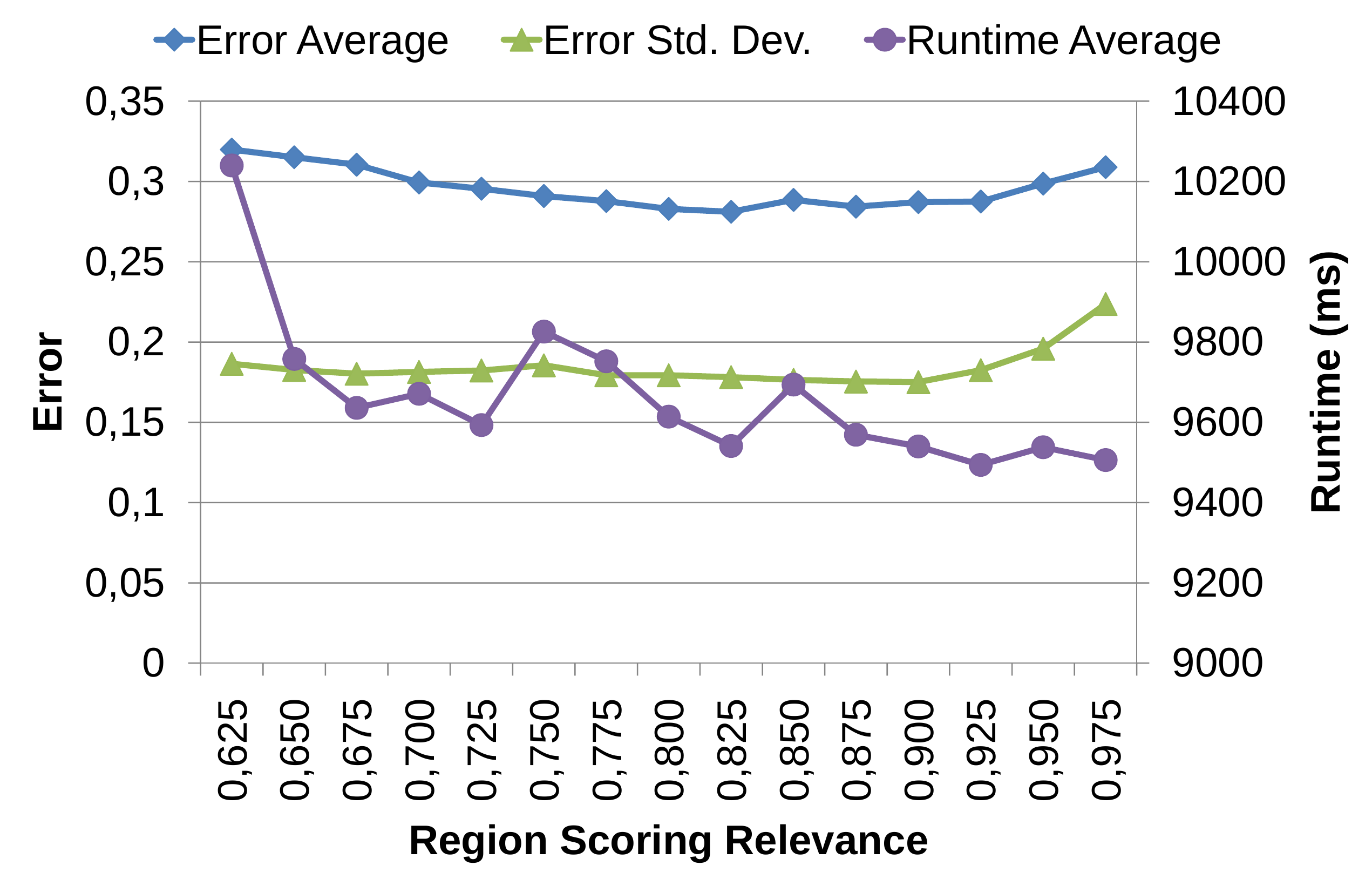}
\par\end{centering}

\caption{\label{fig:tethaRel}Impact of the region scoring parameter $\Theta_{rel}$
on the spatial distortion error and the average runtime per image.}

\end{figure}

In the \emph{region scoring }stage we delete each region \emph{$r$}
from the smoothed approximate mask $I_{app}$ at the \emph{i}-th iteration
if  $MR_{r}^{i}\leq\Theta_{rel}$. Fig.~\ref{fig:tethaRel} shows
the impact of $\Theta_{rel}$ on the segmentation quality.%
Low values of $\Theta_{rel}$ lead to a lower number of deleted regions,
leaving the final mask of the OOI surrounded with a thin border in
most cases. Thus the total amount of false positives is increased
as well as the spatial distortion. The overall influence of $\Theta_{rel}$
is rather low because it only affects the refinement step \emph{Region
Scoring} (see Sec.~\ref{sub:Region-Scoring}).

\section{Impact of DOF on Similarity\label{sec:Similarity}}

Measuring the similarity between two images is an essential step for
content-based image retrieval \cite{zhang2004evaluation} (CBIR).
If the image database contains a significant amount of low DOF images,
a DOF-based segmentation algorithm can improve the classification
accuracy because the extraction of features can be restricted to the
subset of pixels contained in the OOI of the images.

Given for example two images $I_{1}$ and $I_{2}$ which are containing
two semantically different objects $O_{1}$ and $O_{2}$ in their
particularly focussed area in front of a comparatively similar background.
In this case, the distance between image $I_{1}$ and $I_{2}$ should
be significantly lower than between the extracted Objects $O_{1}$
and $O_{2}$, such that $d\left(O_{1},O_{2}\right)\gg d\left(I_{1},I_{2}\right)$. 

\begin{figure*}
\subfloat[\label{fig:imageI1}]{\includegraphics[width=0.23\textwidth]{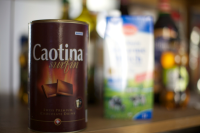}

}\hfill{}\subfloat[\label{fig:ImageI2}]{\includegraphics[width=0.23\textwidth]{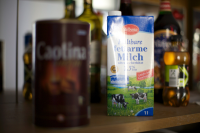}

}\hfill{}\subfloat[\label{fig:ooi1}]{\includegraphics[width=0.23\textwidth]{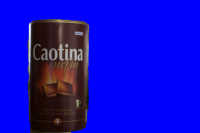}

}\hfill{}\subfloat[\label{fig:ooi2}]{\includegraphics[width=0.23\textwidth]{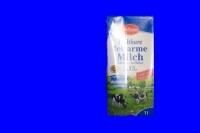}

}\caption{\label{fig:dofSimilarity}Comparing the results of the different algorithms,
where the comparatively similar input images $I_{1}$, $I_{2}$ (Fig.~\ref{fig:imageI1}/\ref{fig:ImageI2})
have homogeneous defocused regions, small DOF and variant colors.
The segmentation results $O_{1}$ and $O_{2}$ (Fig.~\ref{fig:ooi1}
/ \ref{fig:ooi2}) show rather few similarity.}

\end{figure*}

Fig.~\ref{fig:dofSimilarity} shows two sample images that practically
display the same scene with a different distance of the lens to the
focal plane so that different OOIs are displayed sharply. As our main
focus does not lie in the evaluation of best possible feature descriptors
or distance measures, we use the well known color histograms for the
classification of the data set. For each image we thus create a color
histogram \emph{$h$} with 12 bins. Between the color histograms $h\left(I_{1}\right)$
and $h\left(I_{2}\right)$ we can now measure the Minkowski-form Distance
$d_{2}$. For two histograms \emph{$Q$} and \emph{$T$}, $d_{p}$
is defined as in Eq.~\ref{eq:minkowski} which corresponds to the
Euclidean distance in case of $p=2$.\begin{equation}
d_{p}\left(Q,T\right)\text{\textasciiacute=\ensuremath{\left(\overset{N-1}{\underset{i=0}{\sum}}\left(Q_{i}-T_{i}\right)^{p}\right)}}^{\frac{1}{p}}\label{eq:minkowski}\end{equation}
The distance between the color histograms of the complete images $d_{2}\left(h\left(I_{1}\right),h\left(I_{2}\right)\right)=334$
is considerably lower ($3.8$ times) than the distance between the
histograms of their extracted OOIs $d_{2}\left(h\left(O_{1}\right),h\left(O_{2}\right)\right)=1277$
(see Fig.~\ref{fig:ooi1} and Fig.~\ref{fig:ooi2}). This leads
to the assumption, that a CBIR system could profit from an automatic
segmentation of OOIs in low DOF images to improve search quality.

For a brief verification of this hypothesis, we created a database
containing 114 diverse DOF images divided into 17 classes: bird, bee,
cat, coke, deer, eagle, airplane, car, fox, apple, ladybird, lion,
milk, redtulp, yellowtulp and sunflower.

Let $I_{i,j}$ be the \emph{$j$}-th image of the \emph{$i$}-th class
$G_{i}$. We then define the inner-class distance of an image $I_{i,j}$
to be the distance of the image $I_{ij}$ to all other images $I_{ik}$,
$k\neq j$ of the same class $G_{i}$\[
dist_{inner}\left(I_{i,j}\right)=\frac{1}{\left|G_{i}\right|-1}\cdot\underset{J\in G_{i}/I_{i,j}}{\sum}d\left(h\left(J\right),h\left(I_{i,j}\right)\right),\]
where \emph{$d$} is the distance, which is set to the Euclidean Distance
$d_{2}$ in our case. For a given distance measure $d$, the average
inter-class distance $dist_{inter}$ of an image $I_{ij}$ is the
average distance to all other images $J\notin G_{i}$\[
dist_{inter}\left(I_{ij}\right)=\frac{1}{\left|J\in G_{j\neq i}\right|}\cdot\underset{J\in G_{j\neq i}}{\sum}d\left(h\left(J\right),h\left(I_{i,j}\right)\right).\]
In case of a classification task, it is required, that the average
inner-class distance is smaller than the inter-class distance of an
image. To further improve the classification, it is thus desirable
to increase the difference between inter-class and inner-class distance.

In the following experiment, we measured the inner- and inter-class
distance for all classes without any segmentation. Afterwards we applied
the proposed segmentation algorithm to the images, repeated the experiment
and computed the relative changes of the inner- and inter-class distances.

\begin{figure}
\begin{centering}
\includegraphics[width=0.99\columnwidth]{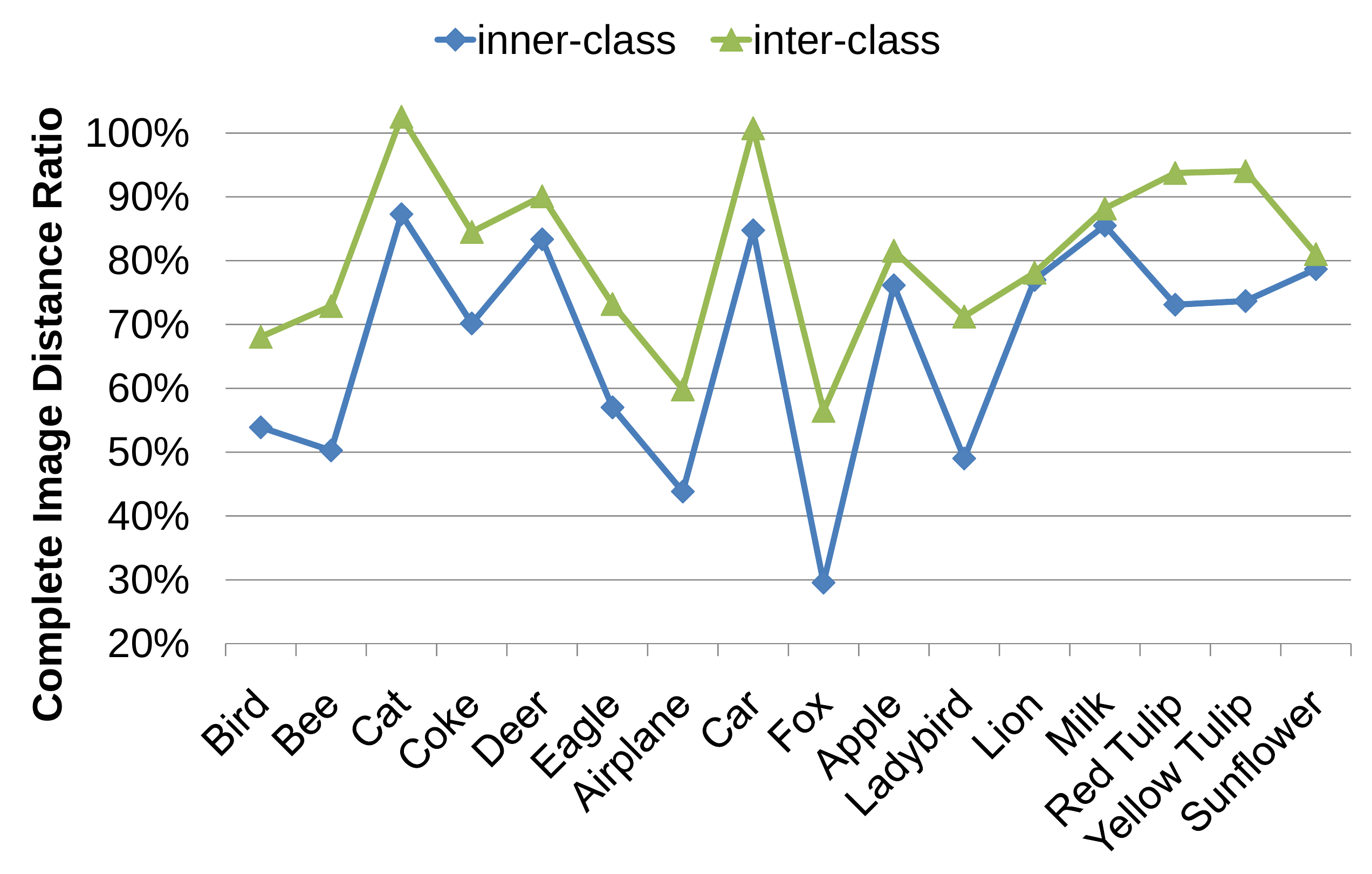}
\par\end{centering}

\caption{\label{fig:similarityChart}Impact of the DOF segmentation on the
inner- and inter-class distance of a test dataset.}

\end{figure}

\begin{figure}
\begin{centering}
\includegraphics[width=0.99\columnwidth]{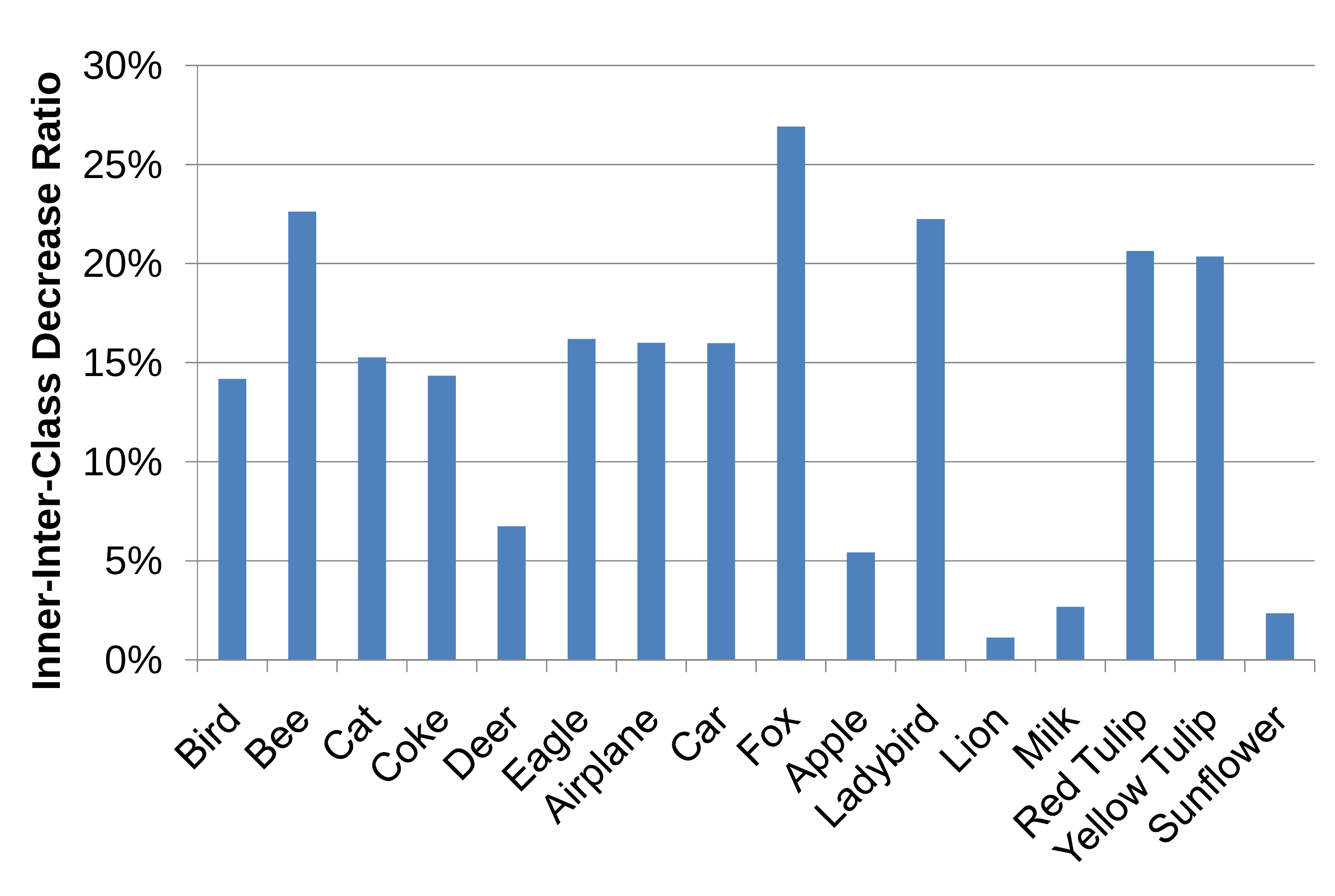}
\par\end{centering}

\caption{\label{fig:similarityRatio}Amount of percent points that the inner-class
distance was lowered more than the inter-class distance.}

\end{figure}

In Fig.~\ref{fig:similarityChart} we summarize the result of the
experiment. %
It can be seen that the inner-class distance is decreased to an average
of $67\%$ of its original value, which is significantly smaller than
the decrease of the inter-class distance which decreases to an average
of $81\%$ of its original value. Thus, the difference between inter-class
and inner-class distance was improved by an average of $14\%$ throughout
the data set with a maximum of $27\%$ in the case of the fox class
and a minimum of $3\%$ in the case of the (glass of) milk class.
These differences between the inner-class distances and the inter-class
distances are illustrated in Fig.~\ref{fig:similarityRatio}. So
it can be said that an CBIR task can profit from the DOF segmentation
if the data set contains enough low DOF images. %

\section{Conclusion\label{sec:Conclusion}}

In this paper a new robust algorithm for the segmentation of low DOF
images is proposed which does not need to set any parameters by hand
as all necessary parameters are determined fully automatically or
preset. Experiments are conducted on diverse sets of real world low
depth of field images from various categories and the algorithm is
compared to three reference algorithms. The experiments show that
the algorithm is more robust than the reference algorithms on all
tested images and that it performs well even if the DOF is growing
larger so that the background begins to show considerable texture.
We also demonstrated the positive impact of low DOF segmentation to
image similarity in case of CBIR. In our future work, we plan to improve
processing speed and accuracy of the algorithm. Furthermore, we plan
to apply the algorithm to movies and to apply an automatic detection,
whether an image is low DOF or not. A Java WebStart demo of the algorithm
can be tested online%
\footnote{http://www.dbs.ifi.lmu.de/research/IJCV-ImageSegmentation/%
}. We also plan to publish the test data as far as image licensing
allows to do so as well as the set including the reference masks for
the ROIs. The implementation of the reference algorithms will be made
available as ImageJ \cite{imagej} plugins for download also at the
demo URL. We also plan to publish the proposed algorithm as an ImageJ
plug-in in the near future.

\section*{Acknowledgements}

This research has been supported in part by the THESEUS program in
the CTC and Medico projects. They are funded by the German Federal
Ministry of Economics and Technology under the grant number 01MQ

\noindent 07020. The responsibility for this publication lies with
the authors.

We would like to thank Philipp Grosselfinger and Sirithana Tiranardvanich
for their ImageJ plug-in implementations of the fuzzy segmentation
\cite{zhang-fuzzy} and video segmentation \cite{li2007unsupervized}.

\bibliographystyle{spmpsci}
\bibliography{references}

\end{document}